\documentclass[]{cit_lab_mfr}

\usepackage{hyperref}
\usepackage{cleveref}
\usepackage{verbatim}

\usepackage{wrapfig}  
\usepackage{graphicx}
\usepackage{subcaption}
\usepackage{listings}
\usepackage{algorithm}
\usepackage{bm}
\usepackage{colortbl}
\usepackage{tocloft} 

\usepackage{xcolor}
\usepackage[dvipsnames]{xcolor}
\usepackage[table]{xcolor}
\usepackage{multirow}
\usepackage{booktabs}
\usepackage{adjustbox}
\usepackage{tabularx}   
\usepackage{siunitx}    
\usepackage{enumitem}
\usepackage[table]{xcolor} 
\usepackage{makecell}
\usepackage{pifont}
\usepackage{textgreek}
\usepackage{xspace}
\usepackage{tikz}
\usepackage{subcaption}
\usepackage{multicol}

\usepackage{tocloft} 

\definecolor{cvprblue}{rgb}{0.21,0.49,0.74}
\definecolor{ForestGreen}{RGB}{34,139,34}
\definecolor{dino}{RGB}{249,231,227}
\definecolor{aliceblue}{rgb}{0.94, 0.97, 1.0}
\definecolor{MyGreen}{HTML}{92D050}
\definecolor{MyBlue}{HTML}{00B0F0}
\definecolor{MyOrange}{HTML}{FFC000}
\definecolor{MyRed}{HTML}{C00000}
\definecolor{myblue}{HTML}{2665B5}
\definecolor{myorange}{HTML}{E68727}
\definecolor{myred}{HTML}{D23D30}
\newcommand{\ours}{\textsc{UniPath}}

\newcommand{\ie}{{\emph{i.e.}}}
\newcommand{\eg}{{\emph{e.g.}}}

\newcommand{\UP}{%
  \tikz[baseline=0ex]\draw[->,line width=1.5pt,green!50!black]
  (2ex,0) -- (2ex,2ex);%
}
\newcommand{\DOWN}{%
  \tikz[baseline=0ex]\draw[<-,line width=1.5pt,red!70!black]
  (2ex,0) -- (2ex,2ex);%
}

\usepackage{subcaption} %

\usepackage[toc,page,header]{appendix}

\usepackage{minitoc}

\title{Beyond Pixel Simulation: Pathology Image Generation via \\ Diagnostic Semantic Tokens and Prototype Control}

\author{%
\parbox{\textwidth}{\centering
Minghao Han$^{1,2,*}$, YiChen Liu$^{3,*}$, Yizhou Liu$^{1,2}$, Zizhi Chen$^{1,2}$ \\ Jingqun Tang$^{4}$, Xuecheng Wu$^{5}$, 
Dingkang Yang$^{1,2,\S}$, Lihua Zhang$^{1,2,\S}$
}}

\affiliation{%
\parbox{\textwidth}{\centering\small
$^1$College of Intelligent Robotics and Advanced Manufacturing, Fudan University\\[1mm]
$^2$Fysics Intelligence Technologies Co., Ltd. (Fysics AI)\\[1mm]
$^3$School of Intelligent Science and Technology, University of Science and Technology Beijing \\[1mm]
$^4$ByteDance \quad
$^5$School of Computer Science and Technology, Xi’an Jiaotong University \\[1mm]

}}

\contribution[*]{Equal contribution}
\contribution[\S]{Corresponding author}

\abstract{
In computational pathology, understanding and generation have evolved along disparate paths: advanced understanding models already exhibit diagnostic-level competence, whereas generative models largely simulate pixels. Progress remains hindered by three coupled factors: the scarcity of large, high-quality image–text corpora; the lack of precise, fine-grained semantic control, which forces reliance on non-semantic cues; and terminological heterogeneity, where diverse phrasings for the same diagnostic concept impede reliable text conditioning. We introduce $\ours$, a semantics-driven pathology image generation framework that leverages mature diagnostic understanding to enable controllable generation. $\ours$ implements Multi-Stream Control: a Raw-Text stream; a High-Level Semantics stream that uses learnable queries to a frozen pathology MLLM to distill paraphrase-robust Diagnostic Semantic Tokens and to expand prompts into diagnosis-aware attribute bundles; and a Prototype stream that affords component-level morphological control via a prototype bank. On the data front, we curate a 2.65M image–text corpus and a finely annotated, high-quality 68K subset to alleviate data scarcity. For a comprehensive assessment, we establish a four-tier evaluation hierarchy tailored to pathology. Extensive experiments demonstrate $\ours$'s SOTA performance, including a Patho‑FID of 80.9 (51\% better than the second-best) and fine-grained semantic control achieving 98.7\% of the real-image. The dataset and code can be obtained from \url{https://github.com/Hanminghao/UniPath}.
}
\date{\today}
\checkdata[Corresponding]{\href{mailto:mhhan22@m.fudan.edu.cn}{mhhan22@m.fudan.edu.cn},\quad \href{mailto:dicken@fysics.ai}{dicken@fysics.ai},\quad \href{mailto:lihuazhang@fudan.edu.cn}{lihuazhang@fudan.edu.cn}}

\begin{document}
\maketitle

\addtocontents{toc}{\protect\setcounter{tocdepth}{-1}}

\section{Introduction}
\label{sec:intro}

Gigapixel Whole Slide Images (WSIs) sit at the core of modern cancer diagnostics and are being reshaped by foundation models~\cite{virchow,gigapath,musk}. Yet pathology AI is evolving along two largely disparate paths: (i) understanding models increasingly capture diagnostic-grade signals across tasks~\cite{cpathomni,pathcopilot,chen2025slidechat, sun2024pathgen}, while (ii) generative models predominantly pursue perceptual realism for augmentation with weak diagnostic conditioning and limited semantic control~\cite{alfasly2025semantic,redekop2025prototype, li2025topofm, pathdiff,yellapragada2025pixcell}. This gap matters as without diagnosis-aware conditioning, generators tend to optimize for appearance cues rather than pathology-grounded semantics.

\begin{wrapfigure}{r}{0.5\textwidth}
  \vspace{-10pt} 
  \begin{center}
    \includegraphics[width=1\linewidth]{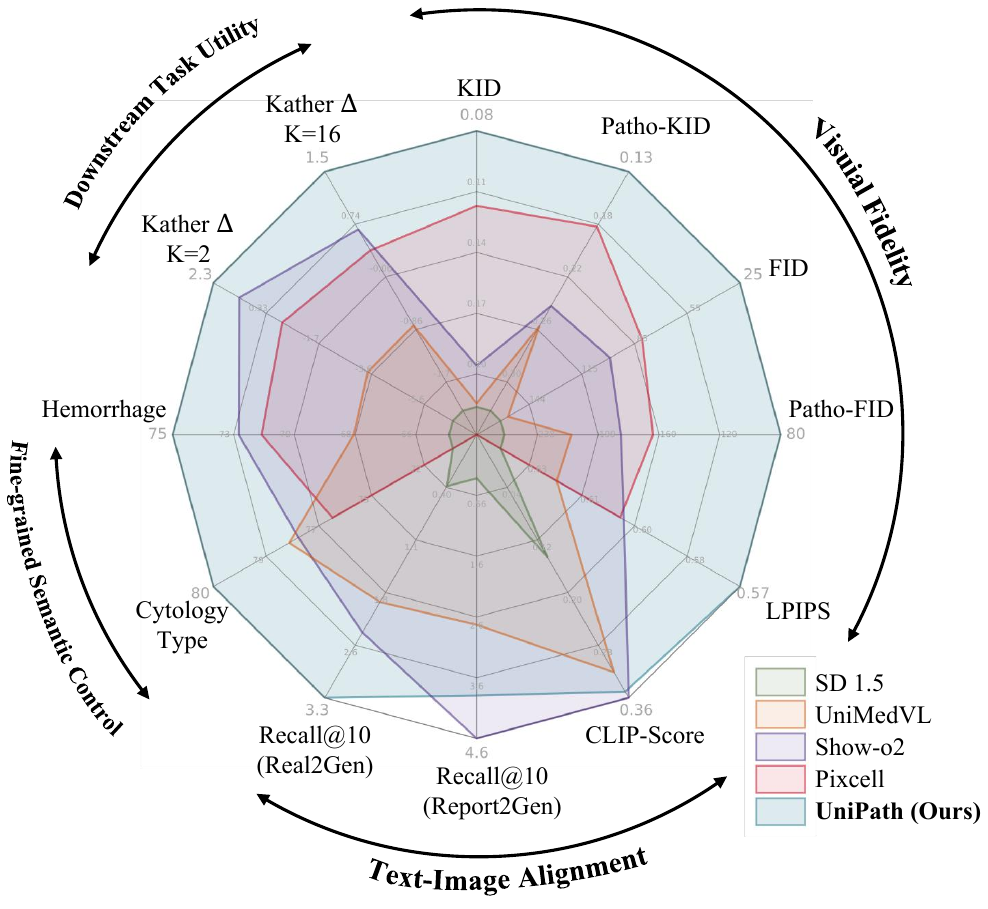}
  \end{center}
  \caption{$\ours$ achieves overall leading performance against state-of-the-art baselines across our four-tier evaluation hierarchy.}
  \label{fig:radar}
  \vspace{-10pt}
\end{wrapfigure}

In practice, progress in pathology text-to-image generation is impeded by three compounding bottlenecks: \textbf{(i) Data scarcity}. Large, high-quality image–text corpora are rare because WSIs are gigapixel-scale and subspecialist annotation is costly, which constrains semantic supervision. \textbf{(ii) Lack of precise semantic control}. Because semantic labels are scarce, SOTA methods rely on non-semantic controls (segmentation masks~\cite{danisetty2025pathsegdiff,xu2025topocellgen} or reference images~\cite{yellapragada2025pixcell}); when present, text-based control is usually restricted to a single cancer with coarse-grained labels~\cite{yellapragada2024pathldm}. \textbf{(iii) Terminological heterogeneity}. The same diagnostic concept is expressed with institution- and pathologist-specific phrasing. General-purpose text encoders~\cite{clip,t5} struggle to align these variants to a consistent meaning, making text conditioning unreliable and weakening semantic control.

We tackle these challenges by unifying diagnostic understanding and image generation in a single model. We present $\ours$, which converts diagnosis-aware semantics into fine-grained, pathology-relevant image generation via \textbf{Multi-Stream Control (MSC)} while retaining strong diagnostic understanding. MSC comprises three streams: \textbf{Raw-Text Stream (RTS)}, which preserves and forwards the user prompt; \textbf{High-Level Semantics Stream (HLS)}, instantiated by learnable queries to a pathology MLLM, distills \textbf{Diagnostic Semantic Tokens (DST)} robust to paraphrase and reporting style and expands surface prompts into diagnosis-aware attribute bundles; and \textbf{Prototype Stream (PS)}, which conditions the generator on retrieved morphology primitives from a prototype bank, enabling component-level control over glandular architecture, nuclear atypia and other key attributes. As a result, $\ours$ unifies diagnostic understanding with semantically controllable generation within a single model, but fully capitalising on this still hinges on abundant and high-quality data.

To this end, we curate a large-scale corpus balancing quantity and diagnostic richness. Starting from 69,044 WSIs in HISTAI~\cite{histai}, rigorous quality control and diversity filtering yield 1.03M diagnostically rich patches. We generate descriptions with a pathology MLLM and merge the resulting pairs with public data to form a 2.65M image–text corpus. For high-fidelity use, we annotate a 68K refined subset with Gemini-2.5 Pro~\cite{comanici2025gemini} and GPT-5~\cite{openai_introducing_gpt5_2025}.

To comprehensively evaluate $\ours$'s controllability and semantic fidelity, we establish a \textbf{Four-Tier Evaluation Hierarchy} tailored for controllable generation in pathology, including \textbf{Visual Fidelity}, \textbf{Text-Image Alignment}, \textbf{Fine-grained Semantic Control}, and \textbf{Downstream Task Utility}. Extensive experiments based on this hierarchy, summarized in Figure~\ref{fig:radar}, demonstrate that $\ours$ achieves SOTA generative fidelity, highly competitive image–text alignment, and exceptional concept-level controllability, while retaining strong diagnostic understanding.
In summary, our main contributions are:
\begin{itemize}[leftmargin=*]
\item We present $\ours$, a unified large multimodal model that couples a pathology understanding module with a controllable generator, enabling semantics-driven pathology image generation while preserving understanding.
\item We design a multi-stream control architecture that combines a high-level semantics stream to mitigate terminological heterogeneity and a prototype stream to enable fine-grained, component-level morphological control.
\item We curate a large-scale, high-quality corpus of 2.65M image–text pairs and a refined 68K subset with pathology-aware quality control to support training and evaluation.
\item We conduct a comprehensive evaluation across fidelity, alignment, controllability, and downstream utility, demonstrating strong image fidelity, robust image–text alignment, and superior concept-level controllability.
\end{itemize}
\section{Related Work}
\label{sec:related_work}

\noindent\textbf{Foundation Models in Pathology.}\hspace{1ex}
Foundation models in pathology are evolving along two disparate paths. On the understanding side, \textbf{Pathology MLLMs} have advanced rapidly, approaching diagnostic-grade capability~\cite{pathcopilot,chen2025slidechat,smartpathr1,wsillava,sun2024pathgen,wsivqa}. Progress is driven by high-quality data, such as Quilt-LLaVA's~\cite{Quilt} instructional narratives, and by novel architectures like CPath-Omni~\cite{cpathomni} (unifying multiple tasks) and Patho-R1~\cite{pathor1} (using RL for reasoning). These works confirm that MLLMs capture complex diagnostic concepts, laying the groundwork for leveraging their strong understanding to guide generation. In contrast, \textbf{Pathology Generative Models} still rely heavily on non-semantic controls to circumvent core challenges. For instance, ToPoFM~\cite{li2025topofm} imposes topology-guided constraints, while others use masks~\cite{pathdiff} or reference images~\cite{yellapragada2025pixcell}. Text-conditioned attempts remain coarse, such as PathLDM~\cite{yellapragada2024pathldm}, which is limited to a single cancer type and provides only coarse-grained textual control. This reliance on general-purpose encoders also fails to normalize pathology phrasing. $\ours$ addresses these gaps via multi-stream control and large-scale, pathology-aware data curation.

\noindent\textbf{Unified Generation and Understanding.}\hspace{1ex}
Unified models in general AI, such as BLIP3o~\cite{blip3o}, BAGEL~\cite{deng2025bagel}, and Show-o2~\cite{showo2}, now couple strong understanding with generation to steer synthesis~\cite{wu2025qwenimage,lin2025uniworld}. Transferring this to pathology is non-trivial, as general-purpose models lack the two key prerequisites for this domain: (i) diagnosis-aware semantics to handle terminological heterogeneity, and (ii) component-level morphology control for specific histological structures. $\ours$ adopts this unified design by introducing multi-stream control to address both challenges: an HLS stream provides diagnosis-aware semantics, while a PS stream enables morphological control.
\section{Data Curation}
\label{sec:data_curation}

To address the data scarcity bottleneck outlined in Section~\ref{sec:intro}, we built two key corpora to support the training and evaluation of $\ours$: (i) a 2.65M-pair large-scale corpus for broad visual-textual alignment, and (ii) a 68K high-quality subset for fine-tuning and evaluation.

\subsection{Large-Scale Pre-training Corpus}
\label{sec:large_scale_data}

Our 2.65M pre-training corpus consists of two components: 1.62M pairs from public data~\cite{sun2024pathgen} and 1.03M high-information patches extracted and annotated from 69,044 HISTAI WSIs~\cite{histai}, thereby increasing data diversity.

\vspace{0.3pt}\noindent\textbf{Representative Patch Selection.}\hspace{1ex}
A dual-strategy pipeline is employed to curate this 1.03M corpus, combining knowledge-guided retrieval and data-driven clustering. First, we use TRIDENT~\cite{TRIDENT} to tile WSIs into $384\times384$ patches at $20\times$ magnification. Features from all patches are then extracted using CONCH~\cite{conch}, and two parallel strategies are applied to balance relevance and diversity:
\begin{itemize}[leftmargin=1.5em]
    \item \textbf{Knowledge-Guided Retrieval:} To capture diagnosis-relevant, information-rich patches, we adopt a unified, knowledge-guided retrieval scheme driven by a powerful LLM. For each WSI, LLM processes metadata in parallel: (i) it uses the diagnosis and organ source to generate organ-specific visual feature descriptions; (ii) it processes the original microscopic examination conclusion. Because this conclusion is often long and exceeds CONCH’s context limit, LLM splits it into multiple short, complete, and independent passages. We then use CONCH to compute the similarity between those queries and all patches, retrieving the clearest, most typical, and most diagnosis-relevant regions.
    \item \textbf{K-means Cluster Sampling:} In parallel, K-means clustering is performed on all patch features. Samples are then drawn from each cluster to ensure morphological diversity and cover the long-tail distribution. The number of clusters depends on the size of the WSI.
\end{itemize}

\vspace{0.3pt}\noindent\textbf{Description Generation.}\hspace{1ex}
Then we de-duplicate patches with UNI2-h~\cite{uni} visual feature similarity $>$ 0.95. Patch-level descriptions were then generated by PathGen-LLaVA~\cite{sun2024pathgen} and subsequently summarized by Qwen3-8B~\cite{qwen3} into refined, information-dense final descriptions. 
WSI tiling and filtering ran for a month using one NVIDIA A800 GPU. The subsequent annotation consumed nearly four days on 16 NVIDIA H100 GPUs.

\subsection{High Quality Refined Subset}
\label{sec:hign_quality_data}

To enable high-fidelity fine-tuning, reliable evaluation, and prototype bank construction, a 68K subset was filtered.

\vspace{0.3pt}\noindent\textbf{Filtering Pipeline.}\hspace{1ex}
Our pipeline first clusters the 2.6M samples via K-means (k=128) on their UNI2-h~\cite{uni} visual features. Subsequently, we compute the Laplacian variance and discard the bottom 50\% with the lowest sharpness. To efficiently filter high-quality, representative samples from it, we employ a two-stage automated sampling strategy: first, we perform proportional random sampling within each cluster to obtain a morphologically diverse and manageable candidate pool. This pool then undergoes an automated quality review, where we use a small MLLM (Qwen3-VL 8B~\cite{qwen3}) to identify and discard all remaining poor-quality samples (\eg, poor staining and fragmented tissue).

\noindent\textbf{Re-annotation \& Partitioning.}\hspace{1ex}
To create high-quality labels, a strict sequential re-annotation process was employed. First, Gemini-2.5 Pro was used to rewrite the descriptions for all candidate samples. Following this, GPT-5 was introduced as an independent reviewer to assess the quality and factual accuracy of each generated description. We only retained pairs that passed the GPT-5 review. From this 68K subset, we first created the 10K Test Set and the 8K Prototype Bank by sampling, prioritizing smaller clusters to cover rare features, and leaving the 50K Fine-tuning Set for Stage 2. Finally, we invited a pathology expert to conduct a spot check: 93.6\% of the image-text pairs were rated as ``Usable'' (detailed in \textbf{Appendix~\ref{sec:appendix_qc}}). We also validated that the Prototype Bank and the Test Set are strictly disjoint, ensuring no data leakage (see \textbf{Appendix~\ref{sec:appendix_risks}}).
\section{Methodology}

\begin{figure}[tp]
    \centering
    \includegraphics[width=1\linewidth]{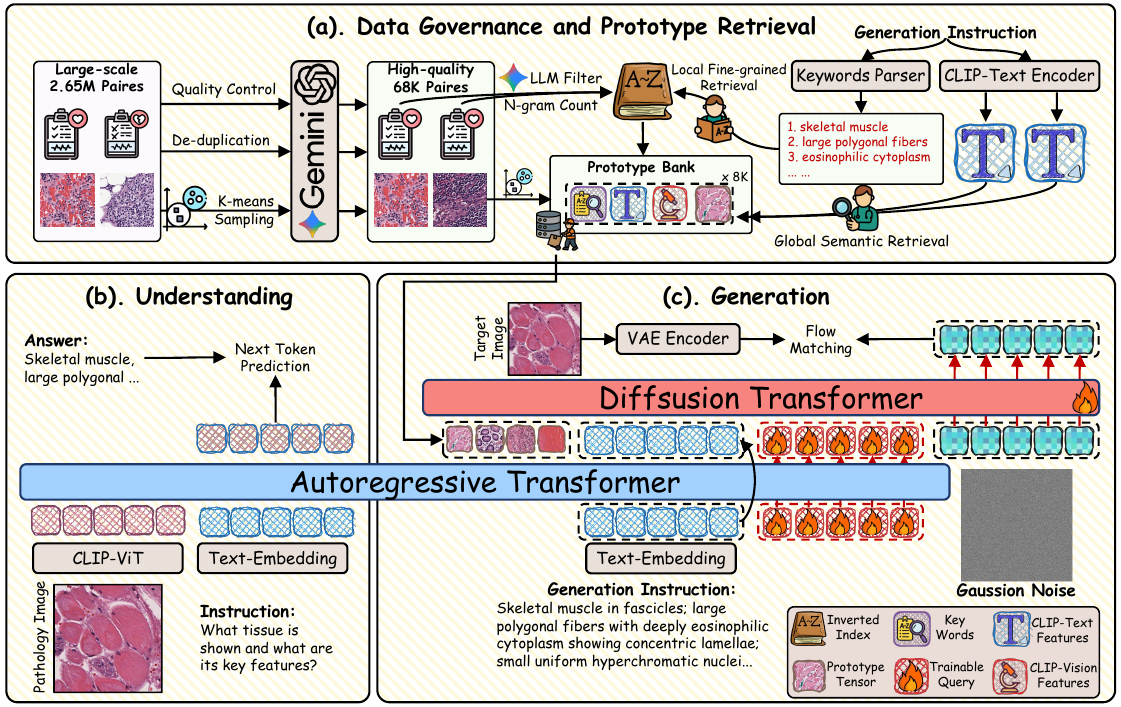}
    \caption{
    \textbf{Overview of $\ours$}: Unifying pathology understanding and synthesis via Multi-Stream Control. \textbf{(a)}: Data governance and prototype retrieval. \textbf{(b)}: A frozen pathology MLLM (Understanding) steers a \textbf{(c)}: generative DiT (Generation) through MSC fusing raw-text, high-level semantics (emitting diagnostic semantic tokens via learnable queries), and prototype cues.
    }
    \label{fig:framework}
\end{figure}

\subsection{Overview of \textbf{$\ours$}}
As illustrated in Figure~\ref{fig:framework}, the $\ours$ architecture is designed as a unified framework that seamlessly integrates diagnostic-grade pathology understanding with high-fidelity, controllable image generation. The framework operates through the synergy of three core components.

\vspace{0.3pt}\noindent\textbf{Understanding Backbone.}\hspace{1ex}
We employ a powerful pathology MLLM (Patho-R1 7B~\cite{pathor1}) as our understanding backbone. Keeping its parameters fully frozen preserves the robust semantic understanding, enabling the model to extract stable, consistent diagnostic semantics and effectively overcome terminological heterogeneity.

\vspace{0.3pt}\noindent\textbf{Generation Backbone.}\hspace{1ex}
We adopt a 0.6B parameter Diffusion Transformer (DiT)~\cite{dit} derived from PixArt-$\alpha$~\cite{chen2023pixart}. This model is more efficient than larger alternatives like SDXL (2.3B)~\cite{podell2023sdxl} and Next-DiT (2B)~\cite{gao2024lumina}, better suited for our domain-specific dataset. We follow the LDM methodology~\cite{ldm}, training the DiT in the VAE's latent space~\cite {sd3}. To achieve higher quality, faster convergence, and more efficient inference, we replace the traditional DDPM objective~\cite{ddpm} with a Flow Matching objective~\cite {lipman2022flow}.

\vspace{0.3pt}\noindent\textbf{The Multi-Stream Control.}\hspace{1ex}
A key innovation of our $\ours$ is the Multi-Stream Control (MSC). The MSC is a trainable module that acts as the interface between the understanding and the generation backbone. Its primary responsibility is to receive and process multi-source control signals originating from the user and the MLLM. It encodes and fuses these signals into a composite conditional sequence, $C_{comp}$. This sequence is then injected into the DiT to steer the generation with fine-grained control.

\vspace{0.3pt}\noindent\textbf{Information Flow.}\hspace{1ex}
During training, the VAE encodes an image to $z_0$ and the MSC encodes its caption to $C_{comp}$. The DiT learns the conditional flow $z_1 \rightarrow z_0$). During inference, the MSC encodes a prompt to $C_{comp}$, guiding the DiT to generate $z_0$. This unified design grants $\ours$ controllability while retaining its understanding capabilities.

\subsection{Multi-Stream Control (MSC)}
\label{sec:MSC}
To handle phrasing heterogeneity and expose component-level morphology control, we propose multi-stream control: it normalizes semantics via a high-level semantics stream, preserves literal intent via a raw-text stream, and injects morphology via a prototype stream.

\vspace{0.3pt}\noindent\textbf{High-Level Semantics (HLS) Stream.}\hspace{1ex}
The HLS stream is designed to both read out phrasing-invariant diagnostic semantics from the frozen pathology MLLM and expand surface prompts into diagnosis-aware attribute bundles, all via a lightweight implementation. We append $N_q$ learnable queries, $Q^{(0)} \in \mathbb{R}^{N_q \times d_c}$, to the end of the prompt embedding sequence $E_{\text{prompt}} \in \mathbb{R}^{L_r \times d_c}$. The combined input sequence $S^{(0)}$ is constructed as:
\begin{small} 
\begin{align}
S^{(0)} = [E_{\text{prompt}} ; Q^{(0)}],
\end{align}
\end{small}%
where $[;]$ denotes concatenation along the sequence dimension. This sequence $S^{(0)}$ is processed by the frozen MLLM backbone. After the final layer $L$, we directly slice the hidden states corresponding to the queries, $H_Q^{(L)} = \mathrm{Tail}_{N_q}(S^{(L)})$, which we term the \textbf{Diagnostic Semantic Tokens (DST)}. This mechanism forces the queries to distill the high-level diagnostic semantics embedded within the user's prompt, without requiring any updates to the backbone parameters.
Finally, these tokens are passed through an LN and an MLP projection to yield the DST condition $C_{\text{DST}}$:
\begin{small} 
\begin{align}
\label{eq:dst_output}
C_{\text{DST}} = \mathrm{MLP}_{\text{DST}}\big(\mathrm{LN}(H_Q^{(L)})\big) \in \mathbb{R}^{N_q \times d_c}.
\end{align}
\end{small}

\vspace{0.3pt}\noindent\textbf{Raw-Text Stream (RTS).}\hspace{1ex}
The RTS stream serves as a complement to the HLS stream by preserving the user's literal intent and textual diversity. While the HLS stream excels at extracting abstract ``consensus semantics,'' it may also over-smooth specific stylistic or nuanced details present in the original prompt. Therefore, we reuse the input embeddings $E_{\text{prompt}} \in \mathbb{R}^{L_r \times d_c}$ and compute the RTS conditional tokens $C_{\text{RTS}}$ via a dedicated $\mathrm{MLP}_{\text{RTS}}$.

\vspace{0.3pt}\noindent\textbf{Prototype Stream (PS).}\hspace{1ex}
The PS stream is designed to achieve component-level morphological control. We employ a non-parametric retrieval mechanism, which remains consistent and frozen during both training and inference. First, we process the user's raw text prompt. We use the powerful pathology vision language model CONCH~\cite{conch} to encode it into an L2-normalized query vector $q \in \mathbb{R}^{d_q}$, which captures the core semantics of the prompt:
\begin{small} 
\begin{align}
q = \operatorname{Norm}\big(\operatorname{CONCH_{text}}(\text{prompt})\big) \in \mathbb{R}^{d_q}.
\end{align}
\end{small}%
To ensure the retrieved prototypes are relevant to the query in terms of both semantic diagnostic concepts and morphological visual appearances, we adopt a hybrid retrieval strategy: (i) \textbf{Global Semantic Retrieval}: We use $q$ to perform Top-k cosine similarity retrieval on both the text index $I_{\text{text}}$ and the vision index $I_{\text{vision}}$:
\begin{small} 
\begin{align}
U_g = \operatorname{TopK}_{k_t}(I_{\text{text}} \cdot q) \cup \operatorname{TopK}_{k_v}(I_{\text{vision}} \cdot q).
\end{align}
\end{small}%
(ii) \textbf{Local Fine-grained Retrieval}: We parse keywords $\mathcal{K}$ from $q$ and query an inverted index $\mathcal{I}$ to recall fine-grained morphological keywords, yielding the set $U_{l} = \bigcup_{w \in \mathcal{K}} \mathcal{I}(w)$.
We then take the union of prototype IDs recalled from both strategies, clip this sequence to a fixed length $K_{m}$, and retrieve the corresponding features:
\begin{small} 
\begin{align}
P = \mathrm{Proto}[\hat{U}] \in \mathbb{R}^{K_{m} \times d}, \quad
\hat{U} = \operatorname{Clip}\big( (U_g \cup U_{l}), K_{m} \big),
\end{align}
\end{small}%
where $\mathrm{Proto} \in \mathbb{R}^{M \times d_p}$ is our offline bank storing the UNI2-h~\cite{uni} features for $M$ prototypes. We also compute the prototype conditional tokens $C_{\text{PS}}$ via $\mathrm{MLP}_{\text{PS}}$.
Finally, the three conditional tokens are fused into a single composite sequence via concatenation:
\begin{small}
\begin{align}
C_{comp} = [C_{\text{DST}} ; C_{\text{RTS}} ; C_{\text{PS}}].
\end{align}
\end{small}%
This $C_{comp}$ is injected into the DiT via cross-attention.

\subsection{Prototype Bank Construction}
\label{sec:prototype}

The Prototype Bank is a non-parametric, frozen instance bank of 8K real samples (rather than K-means centroids) to preserve true morphological diversity.

\vspace{0.3pt}\noindent\textbf{Retrieval Indices.}\hspace{1ex}
We built three index components to support our hybrid retrieval strategy:
\begin{itemize}[leftmargin=1.5em]
    \item \textbf{Dense Indices ($I_{\text{text}}$ \& $I_{\text{vision}}$):} We built two L2-normalized dense indices, $I_{\text{text}}$ and $I_{\text{vision}}$ ($\in \mathbb{R}^{M \times d_q}$), by encoding the 8K refined texts and images using the CONCH text and vision encoders, respectively.
    \item \textbf{Inverted Index ($\mathcal{I}$):} We created a pathology vocabulary by extracting Top-5000 N-gram candidates from the 50K subset, and then using Gemini-2.5 Pro to review and refine them based on pathology rules. This vocabulary was then used to parse the 8K prototype bank texts to build the inverted index $\mathcal{I}$.
\end{itemize}

\vspace{0.3pt}\noindent\textbf{Prototype Feature Bank.}\hspace{1ex}
The Prototype Feature Bank ($\mathrm{Proto}$) provides the content for injection. It was generated by extracting features from the 8K images using the UNI2-h extractor, resulting in the final matrix $\mathrm{Proto} \in \mathbb{R}^{M \times d_p}$.

\subsection{Training Strategy}
\label{sec:training}

\vspace{0.3pt}\noindent\textbf{Flow Matching Objective.}\hspace{1ex}
We adopt the Flow Matching (FM)~\cite{lipman2022flow} objective to train our DiT backbone. The FM framework learns a vector field to transport samples from a prior distribution (\eg, Gaussian) to a target continuous distribution.
Given a ground-truth latent $z_0$ (from VAE encoder) and our composite condition $C_{\text{comp}}$, the training proceeds as follows. At each step, we sample a timestep $t \sim \mathcal{U}(0, 1)$, and noise $z_1 \sim \mathcal{N}(0, 1)$. Following~\cite{liu2022flow}, we compute the interpolated latent $z_t$ via linear interpolation:
\begin{small}
\begin{align}
\label{eq:fm_interp}
z_t = t z_0 + (1 - t) z_1. 
\end{align}
\end{small}%
\noindent The corresponding target velocity vector $v_t$ (\ie, the derivative of $z_t$ with respect to $t$) is analytically given by:
\begin{small}
\begin{align}
\label{eq:fm_velocity}
v_t = \frac{d z_t}{dt} = z_0 - z_1.
\end{align}
\end{small}%
Our DiT model, parameterized by $\theta$ and denoted as $v_\theta$, is trained to predict this velocity, conditioned on the corrupted latent $z_t$, the timestep $t$, and our composite condition $C_{\text{comp}}$. The training objective $\mathcal{L}_{\text{Flow}}$ is defined as the L2 loss:
\begin{small}
\begin{align}
\label{eq:fm_loss}
\mathcal{L}_{\text{Flow}}(\theta) = \mathbb{E}_{z_0, C_{\text{comp}}, t, z_1} \left[ \left\| v_\theta(z_t, t, C_{\text{comp}}) - v_t \right\|^2 \right]. 
\end{align}
\end{small}

\vspace{0.3pt}\noindent\textbf{Two-Stage Training Strategy.}\hspace{1ex}
We employ a two-stage training strategy. Stage 1 (Semantic Alignment) uses the 2.65M large-scale corpus to pre-train the DiT ($v_\theta$) and MSC modules to learn fundamental visual-textual alignment. Stage 2 (High-Quality Fine-tuning) then continues to train the DiT and MSC on the 50K high-quality fine-tuning set using a smaller learning rate, significantly enhancing visual fidelity and fine-grained controllability.

{\renewcommand{\arraystretch}{1.2}
\begin{table*}[tp]
\centering
\caption{Quantitative comparison of \textbf{Visual Fidelity} and \textbf{Text-Image Alignment} on our 10K High-Quality Test Set. Patho-FID/KID are computed using the UNI2-h backbone. \textsuperscript{\textcolor{orange}{\scalebox{1.2}{\ding{72}}}} indicates models fully fine-tuned on our large dataset. Retrieval metrics (Recall/mAP) are reported as Report2Gen (T2I) / Real2Gen (I2I). The best-performing model is \textbf{in-bold}, and the second-best-performing model is \underline{underlined}.}
\label{tab:tier1&2}
\resizebox{\linewidth}{!}{
\begin{tabular}{lccccccccccc} 
\toprule
\multirow{2}{*}{\textbf{}} &\multirow{2}{*}{\textbf{\makecell{Unif.\\Model}}} & \multicolumn{5}{c|}{\textbf{Visual Fidelity}~~$\DOWN$}  & \multicolumn{5}{c}{\textbf{Text-Image Alignment}~~$\UP$}                     \\ 
\cline{3-12}
&  \rule{0pt}{3ex}  & FID    & KID   & Patho-FID & Patho-KID & LPIPS & CLIP-Score & Recall@10 & ~Recall@50  & mAP@10    & mAP@50     \\ 
\hline
\rowcolor{dino}
\multicolumn{1}{l}{\textcolor{gray}{Real Data}} &
\textcolor{gray}{-} & \textcolor{gray}{-} & \textcolor{gray}{-} & \textcolor{gray}{-} &
\textcolor{gray}{-} & \textcolor{gray}{-} &
\textcolor{gray}{0.372} & \textcolor{gray}{9.60/-} &
\textcolor{gray}{27.55/-} & \textcolor{gray}{3.45/-} & \textcolor{gray}{4.21/-} \\
\hline
\multicolumn{12}{c}{\textbf{\textit{General Text to Image Generation Models}}}  \\ 
\hline
SD1.5\textsuperscript{\textcolor{orange}{\scalebox{1.2}{\ding{72}}}}~\cite{ldm}   & \textcolor{red!80!black}{\ding{56}}                            & 160.36 & 0.221 & 259.69    & 0.321     & 0.634 & 0.147      & 0.27/0.38 & 1.48/1.33   & 0.23/0.49 & 0.40/0.61  \\
SDXL\textsuperscript{\textcolor{orange}{\scalebox{1.2}{\ding{72}}}}~\cite{podell2023sdxl}  & \textcolor{red!80!black}{\ding{56}}                            & 291.17 & 0.305 & 295.50    & 0.343     & 0.657 & -0.035     & 0.34/0.19 & 1.21/0.95   & 0.52/0.27 & 0.67/0.42  \\
Pixart-$\alpha$\textsuperscript{\textcolor{orange}{\scalebox{1.2}{\ding{72}}}}~\cite{chen2023pixart}   & \textcolor{red!80!black}{\ding{56}}                            & 270.36 & 0.277 & 336.78    & 0.401     & 0.603 & 0.105      & 0.84/0.64 & 3.40/2.75   & 1.17/0.91 & 1.38/1.15  \\
BLIP3o\textsuperscript{\textcolor{orange}{\scalebox{1.2}{\ding{72}}}}~\cite{blip3o}  & \textcolor{green!60!black}{\ding{52}}                            & 95.04  & 0.183 & 358.61    & 0.431     & 0.598 & 0.080      & 0.93/1.10 & 3.78/3.68   & 1.23/1.61 & 1.51/1.72  \\
Show-o2\textsuperscript{\textcolor{orange}{\scalebox{1.2}{\ding{72}}}}~\cite{showo2}   & \textcolor{green!60!black}{\ding{52}}                            & 98.82  & 0.200 & 184.19    & 0.239     & \underline{0.601} & \textbf{0.357}      & \textbf{4.64}/\underline{2.40} & \textbf{15.26}/\underline{8.15}  & \textbf{5.85}/\underline{2.74} & \underline{5.50}/2.58  \\ 
\hline
\multicolumn{12}{c}{\textbf{\textit{Pathological / Medical Text to Image Generation Models}}}  \\ 
\hline
Pixcell~\cite{yellapragada2025pixcell}    & \textcolor{red!80!black}{\ding{56}}                            & \underline{80.74}  & \underline{0.119} & \underline{163.44} & \underline{0.177}     & 0.602 & -          & -         & ~-          & ~-        & -          \\
PathLDM~\cite{yellapragada2024pathldm}   & \textcolor{red!80!black}{\ding{56}}                            & 93.91 & 0.170 & 174.32    & 0.254     & 0.606 & 0.182      & 0.12/0.19          & 0.70/0.66  &  0.19/0.27 & 0.30/0.36   \\
UniMedVL~\cite{ning2025unimedvl}                & \textcolor{green!60!black}{\ding{52}}                            & 156.17 & 0.219 & 216.27    & 0.255     & 0.619 & 0.319      & 2.73/1.98 & 8.47/6.67   & 3.86/2.62 & 3.85/\underline{2.60}  \\
\rowcolor{aliceblue} \textbf{$\ours$ (Ours)}           & \textcolor{green!60!black}{\ding{52}}                            & \textbf{25.70}  & \textbf{0.081} & \textbf{80.86}     & \textbf{0.134}  & \textbf{0.570} & \underline{0.348}      & \underline{3.92}/\textbf{3.30} & \underline{14.08}/\textbf{11.92} & \underline{5.66}/\textbf{4.64} & \textbf{5.54}/\textbf{4.67}  \\
\bottomrule
\end{tabular}
}
\end{table*}}
\section{Experiments and Results}
\label{sec:expermiments}

\subsection{Implementation Details}
\label{sec:implementation}

\vspace{0.3pt}\noindent\textbf{Model Architecture.}\hspace{1ex}
Our DiT backbone (0.6B) comprises 28 Transformer layers, 16 heads, and a hidden dimension of 1152. We use the Stable Diffusion 3 VAE~\cite{sd3} with 8$\times$ downsampling. The understanding backbone (Patho-R1 7B)~\cite{pathor1} and the VAE are frozen during training. We use 64 learnable queries in the HLS stream, yielding 64 DST.

\vspace{0.3pt}\noindent\textbf{Training Details.}\hspace{1ex}
In Stage 1, we utilize 2.58M text-image pairs (excluding the 68K subset) and train for 10,000 steps with a global batch size of 512. The learning rate is linearly warmed up to a peak of $1e^{-4}$, followed by cosine annealing. Stage 2 loads the pretrained weights and fine-tunes exclusively on the 50K data for 500 steps with a fixed learning rate of $2e^{-5}$. For the PS Stream, we set $K_m=16$. All experiments are conducted on 16 NVIDIA H100 GPUs.

\subsection{Evaluation Protocols}
\label{sec:protocols}

\vspace{0.3pt}\noindent\textbf{Four-Tier Evaluation Hierarch.}\hspace{1ex}
Traditional metrics are insufficient for evaluating pathology generative models, as they fail to capture diagnostic relevance or controllability. We thus first performed a preliminary validation to confirm the model's understanding, then established a four-tier evaluation scheme.
\textbf{Tier 1: Visual Fidelity.}
We assess perceptual quality and distributional fidelity using FID~\cite{fid} and KID~\cite{kid} (standard and pathology variants), as well as LPIPS~\cite{LPIPS}.
\textbf{Tier 2: Text-Image Alignment.}
We assess semantic consistency using clip-score, retrieval metrics, and MLLM, as well as human judges.
\textbf{Tier 3: Fine-grained Semantic Control.}
We assess fine-grained semantic control via the ``Train-on-Synth, Test-on-Real'' paradigm, comparing the real-test-set performance of a Gen2Real classifier against a Real2Real classifier.
\textbf{Tier 4: Downstream Task Utility.}
We evaluate data augmentation utility by comparing the real-test-set performance gain of a baseline classifier (real) versus an augmented classifier (real + synthetic).

\vspace{2pt}\noindent\textbf{Baselines.}\hspace{1ex}
To comprehensively evaluate $\ours$'s generative capabilities, we selected SOTA models from both general and specific domains as baselines.
\textbf{General T2I} SOTA:
We selected SD1.5~\cite{ldm}, SDXL~\cite{podell2023sdxl}, PixArt-$\alpha$~\cite{chen2023pixart}, BLIP3o~\cite{blip3o}, and Show-o2~\cite{showo2}. As these general-purpose models have minimal exposure to pathology data, we fully fine-tuned them on our dataset using the same training strategy.
\textbf{Pathology-Specific} SOTA:
We selected PixCell~\cite{yellapragada2025pixcell}, PathLDM~\cite{yellapragada2024pathldm}, and UniMedVL~\cite{ning2025unimedvl}, which are leading generative models designed for this domain. Notably, PixCell is natively image-conditioned. To adapt PixCell for T2I comparison , we used CONCH to retrieve a prompt-relevant image from our prototype bank to serve as its generation condition. We note that this adaptation, while necessary, is a convenience variant and may not represent the optimal text-conditioning performance of the PixCell architecture.

\vspace{2pt}\noindent\textbf{Understanding Capability.}\hspace{1ex}
For validation, we evaluate $\ours$ on the PathMMU benchmark~\cite{sun2024pathmmu}. The results (see \textbf{Appendix~\ref{sec:appendix_tier0}}) show that $\ours$ achieves SOTA performance among open-source models and approaches closed-source models. This performance confirms the model's ability to extract robust, phrasing-invariant semantics, which are critical to controllable generation.

\subsection{Visual Fidelity}
\label{sec:visual_fidelity}

\begin{figure}[t]
  \centering
  \begin{subfigure}[t]{0.45\linewidth}
    \centering
    \includegraphics[width=\linewidth]{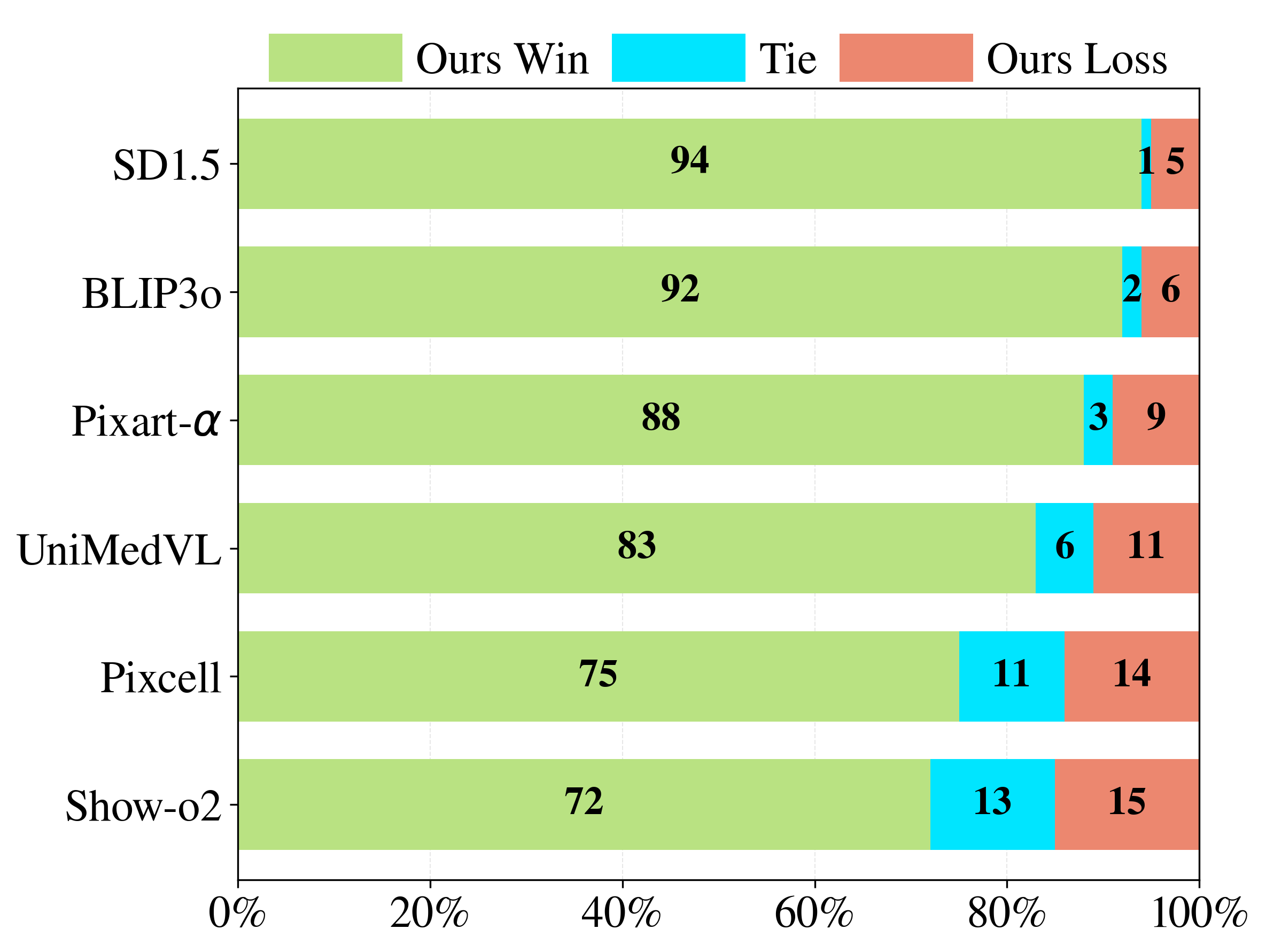}
    \caption{GPT-5 as judge.}
    \label{fig:llm_as_judge:b}
  \end{subfigure}
  \begin{subfigure}[t]{0.45\linewidth}
    \centering
    \includegraphics[width=\linewidth]{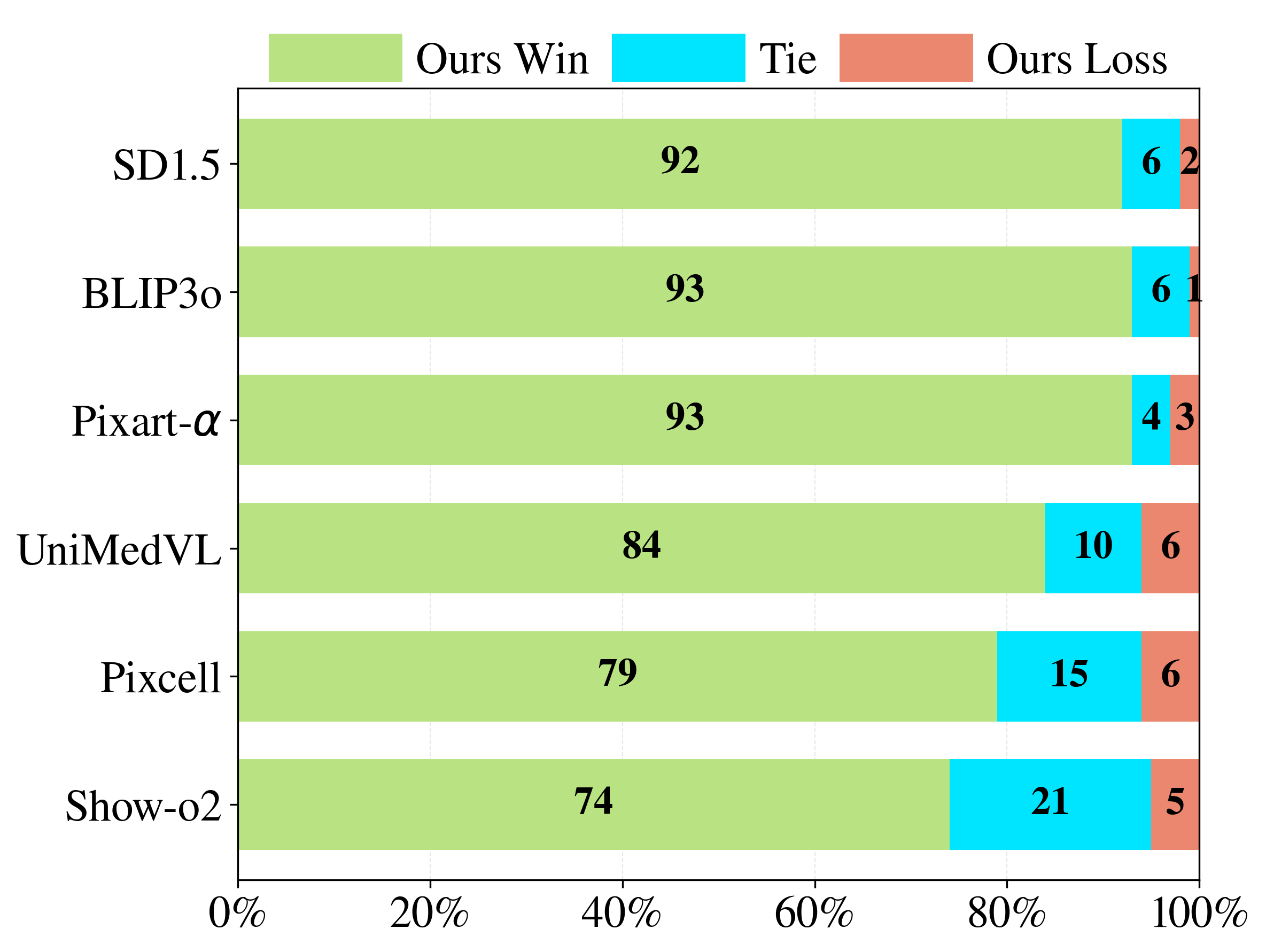}
    \caption{Human experts as judges.}
    \label{fig:llm_as_judge:b}
  \end{subfigure}
  \caption{MLLM and human expert evaluation results.}
  \label{fig:llm_as_judge}
\end{figure}

\begin{wraptable}{r}{0.56\textwidth}
\centering
{\renewcommand{\arraystretch}{1.1}
\resizebox{\linewidth}{!}{
\begin{tabular}{lcccc} 
\toprule
\multirow{2}{*}{Models} & \multicolumn{2}{c}{\textbf{Cytology Type (4-classes)}} & \multicolumn{2}{c}{\textbf{Hemorrhage (2-classes)}}  \\ 
\cline{2-5}
                        & Wgt. F1 & Wgt. AUC         & Wgt. F1 & Wgt. AUC      \\ 
\hline
\rowcolor{dino}
\multicolumn{1}{l}{\textcolor{gray}{Real Data-Image}} & \textcolor{gray}{83.43} &\textcolor{gray}{87.15} &\textcolor{gray}{78.13} &\textcolor{gray}{80.11} \\
\rowcolor{dino}\multicolumn{1}{l}{\textcolor{gray}{Real Data-Text}} &\textcolor{gray}{95.22} &\textcolor{gray}{98.59} &\textcolor{gray}{83.09} &\textcolor{gray}{88.29} \\

\hline
SD1.5\textsuperscript{\textcolor{orange}{\scalebox{1.2}{\ding{72}}}}~\cite{ldm} & 71.26     & 71.34     & 64.41     & 62.88           \\
SDXL\textsuperscript{\textcolor{orange}{\scalebox{1.2}{\ding{72}}}}~\cite{podell2023sdxl}   & 69.41     & 42.31     & 62.25     & 56.78           \\
Pixart-$\alpha$\textsuperscript{\textcolor{orange}{\scalebox{1.2}{\ding{72}}}}~\cite{chen2023pixart} & 70.65     & 72.27    & 67.67     & 68.53           \\
BLIP3o\textsuperscript{\textcolor{orange}{\scalebox{1.2}{\ding{72}}}}~\cite{blip3o}                  & 67.31     & 68.78              & 68.57     & 65.38           \\
Show-o2\textsuperscript{\textcolor{orange}{\scalebox{1.2}{\ding{72}}}}~\cite{showo2}                 & 77.82     & 66.89              & 72.44     & 71.72           \\ 

\hline
Pixcell~\cite{yellapragada2025pixcell}                 & 76.37     & 83.22              & 71.57     & 71.59           \\
PathLDM~\cite{yellapragada2024pathldm}   & 69.57  &63.61&62.07&47.58  \\
UniMedVL~\cite{ning2025unimedvl}                & 78.21     & 72.43              & 68.05     & 65.36           \\
\hline
\rowcolor{aliceblue} \textbf{$\ours$ (Ours)}         & \underline{81.43}     & \underline{84.05}      & \underline{74.96}     & \underline{75.27}           \\
\rowcolor{aliceblue} \textbf{$\ours$-\textit{Aug.}}   & \textbf{81.49}     & \textbf{85.29}              & \textbf{77.02}     & \textbf{79.03}           \\
\bottomrule
\end{tabular}
}
}
\caption{Results on \textbf{Fine-grained Control}.}
\label{tab:tier3}
\end{wraptable}

We assess visual fidelity in Table~\ref{tab:tier1&2} (left), covering standard (FID/KID) and pathology (Patho-FID/KID with UNI2-h~\cite{uni} backbone) metrics, as well as LPIPS. $\ours$ attains state-of-the-art results across all metrics. The gap is substantial: on the challenging Patho-FID, $\ours$ achieves a 50.5\% relative reduction compared with the second-best model. It also yields the best scores on FID, KID, Patho-KID, and LPIPS, indicating that the images generated by $\ours$ align best with the real data distribution in both general and pathology feature spaces. We also show results using Virchow2~\cite{zimmermann2024virchow2} and MUSK~\cite{musk} extractors in \textbf{Appendix~\ref{sec:appendix_more_metrics}}, where performance remains strong.

\subsection{Text-Image Alignment}
\label{sec:text_image_align}

We next evaluate the models' performance on text-image alignment. Table~\ref{tab:tier1&2} (right) reports the quantitative results based on a CONCH-based clip-score and retrieval metrics.

\vspace{2pt}\noindent\textbf{CLIP-Score.}\hspace{1ex}
On the CLIP-Score metric, our $\ours$ achieves a high score of 0.348, beating all non-unified models. It trails only the unified model Show-o2 (0.357), confirming that a strong understanding aids semantic generation. The 0.009 CLIP‑Score gap to Show‑o2 is explained by its SigLIP‑distilled semantic tokens (see \textbf{Appendix~\ref{sec:appendix_limitations}}); other alignment and downstream metrics favour $\ours$.

\vspace{2pt}\noindent\textbf{Retrieval Metrics.}\hspace{1ex}
On Report2Gen (T2I retrieval), Show-o2 achieves the best results, while $\ours$ is highly competitive with near-tied mAP. On Real2Gen (I2I retrieval)—which measures feature-space distance to real images—$\ours$ reaches clear SOTA, outperforming all methods by a large margin. Thus, although Show-o2 has a slight T2I edge, $\ours$ generates images that are closer to real pathology in feature space. This trend matches the LPIPS results and shows that $\ours$ is not only semantically aligned but also produces morphology and visual structures most faithful to the true pathological appearance.

\begin{figure}[t]
  \centering
  \begin{subfigure}[t]{0.4\linewidth}
    \centering
    \includegraphics[width=\linewidth]{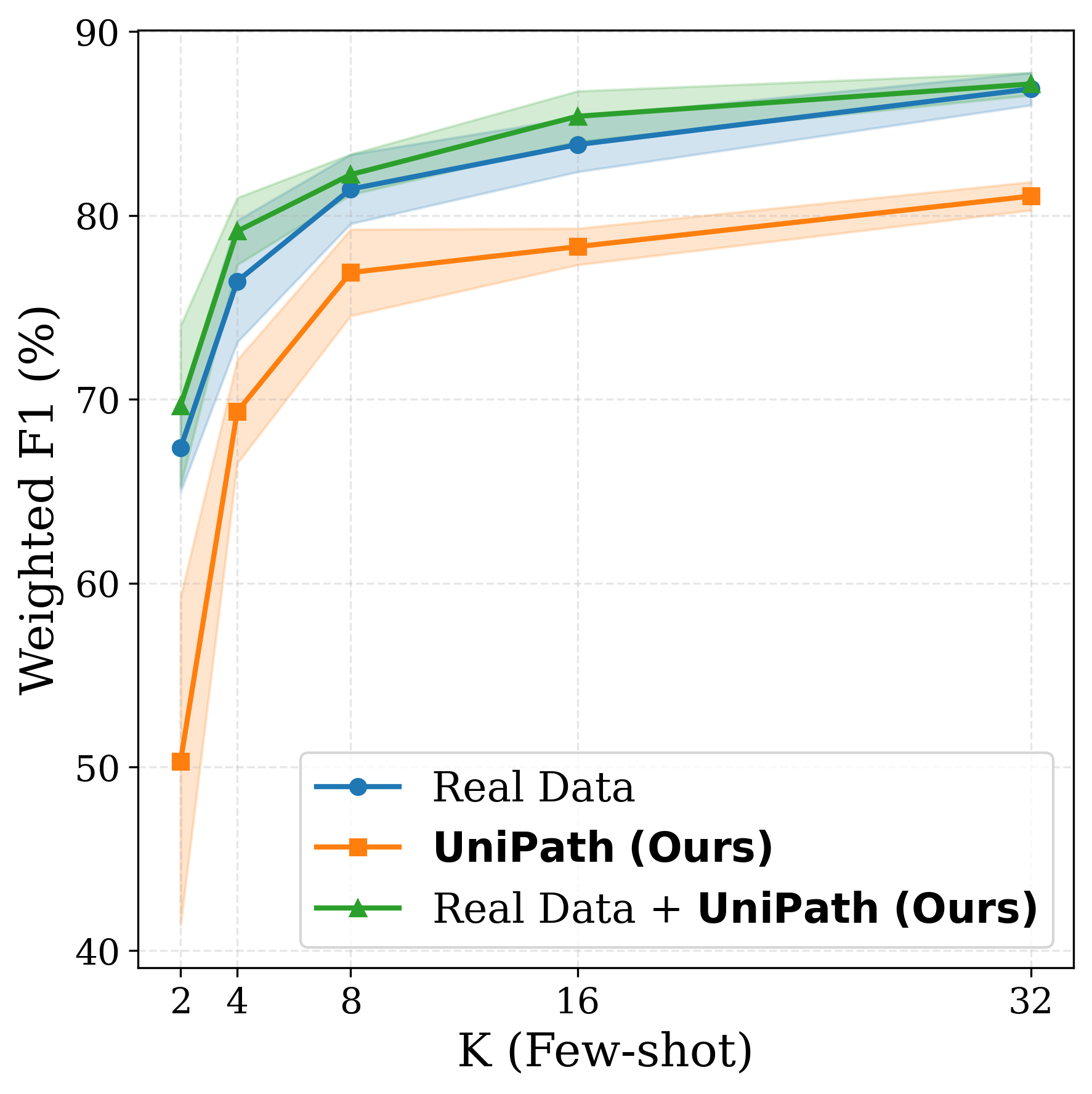}
    \caption{Gen data boosts downstream F1.}
    \label{fig:tier4:a}
  \end{subfigure}
  \begin{subfigure}[t]{0.4\linewidth}
    \centering
    \includegraphics[width=\linewidth]{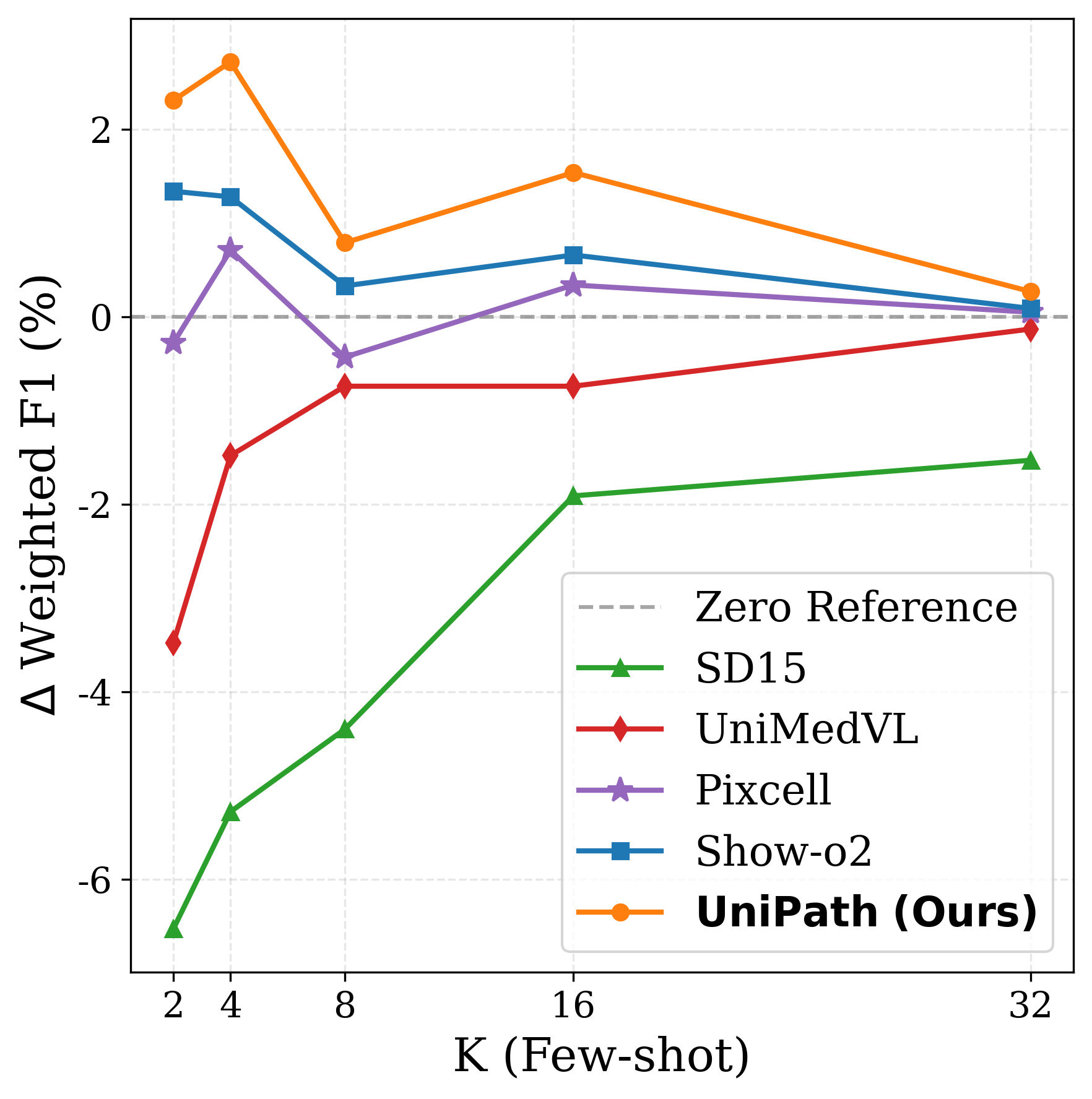}
    \caption{$\Delta$F1 of \ours\ over baselines.}
    \label{fig:tier4:b}
  \end{subfigure}
  \caption{Few-shot classification on Kather-CRC-2016.}
  \label{fig:tier4}
\end{figure}

\vspace{2pt}\noindent\textbf{As-Judge Evaluation.}\hspace{1ex}
This strong alignment is confirmed by two pairwise evaluations (Figure~\ref{fig:llm_as_judge}). Against the strongest baseline, $\ours$ was preferred by GPT-5 in 72\% of cases. A panel of three human pathologists confirmed this, preferring $\ours$ in 74\% of cases. This convergence of both MLLM and human expert preference underscores a clear advantage in nuanced, human-aligned semantic understanding. See details in \textbf{Appendix~\ref{sec:appendix_as_judge}}.

\subsection{Fine-grained Semantic Control}
\label{sec:fine_grained}

We assess semantic fidelity for fine-grained concepts with a ``Train-on-Synth, Test-on-Real'' (G2R) protocol, since FID does not ensure learnable morphology; specifically, we split the 10K test set 60/20/20 (train/val/test), train a linear probe classifier on frozen CONCH features, and evaluate on the real test split, reporting F1 and AUC.

\vspace{2pt}\noindent\textbf{Quantitative Analysis.}\hspace{1ex}
Table~\ref{tab:tier3} summarizes the results. Real Data–Image serves as the Real2Real (R2R) ceiling, while the strong Real Data–Text baseline indicates that key diagnostic factors are encoded in the prompts. Among all G2R models, $\ours$ achieves performance most comparable to the R2R benchmark. For cytology type, its F1 score is within two points of the R2R result, and for hemorrhage, it surpasses the next-best model by a clear margin.

\vspace{2pt}\noindent\textbf{Closing the Gap with Augmentation.}\hspace{1ex}
We also evaluated augmenting with 5 images per prompt ($\ours$-Aug), which further boosts performance, reaching 98.7\% (hemorrhage) and 97.9\% (cytology type) of the real-image AUC. This nearly closes the R2R gap, indicating that the synthetic features are morphologically precise enough to be learned and generalized to real images.

\subsection{Downstream Task Utility}
\label{sec:fine_grained}

Finally, we assess $\ours$'s practical utility as a data augmentation tool using a few-shot setting.

\begin{figure}[t]
  \centering
  \includegraphics[width=0.8\linewidth]{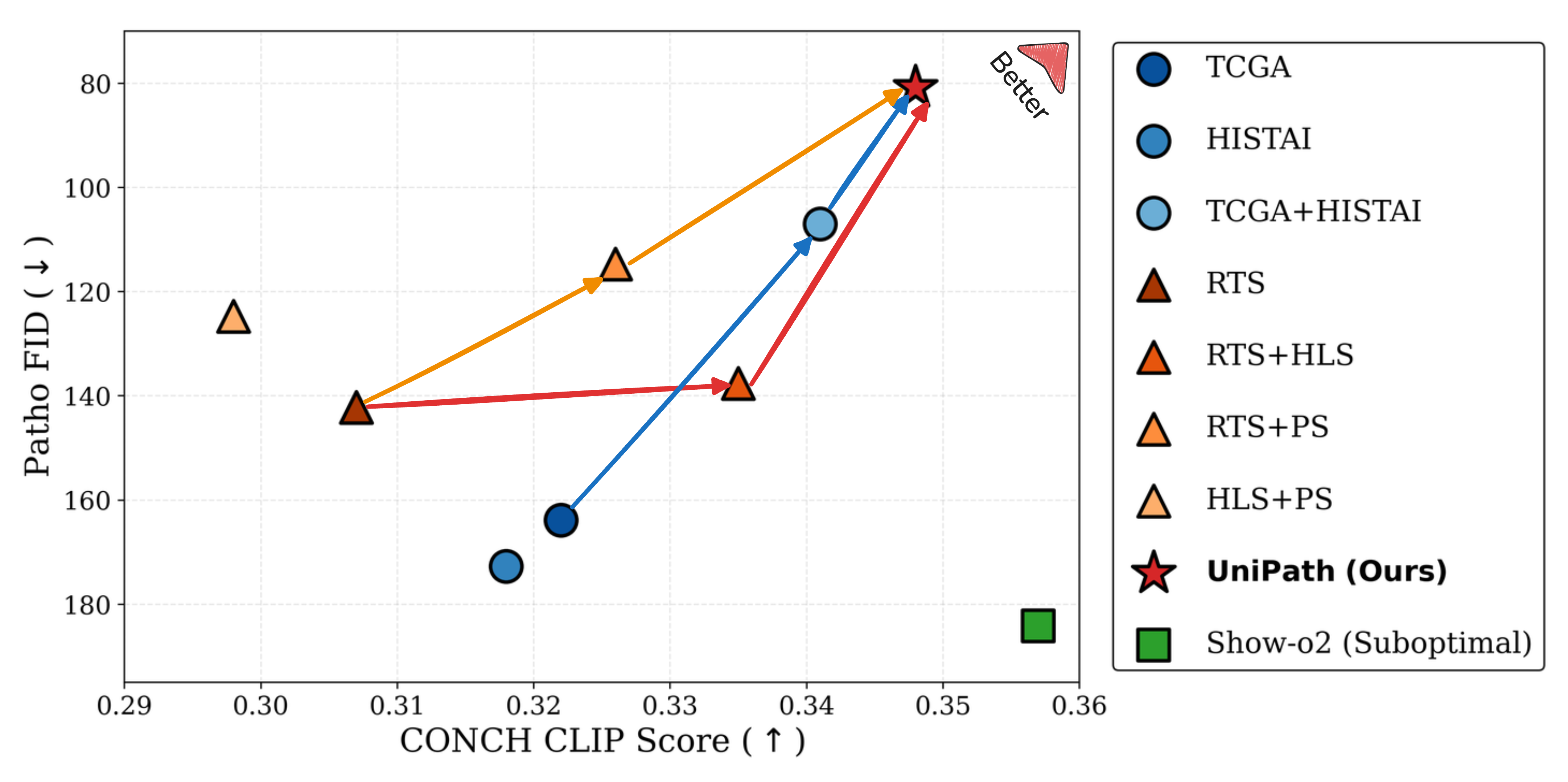}
  \caption{Dataset and component ablation results.}
  \label{fig:ablation}
\end{figure}

\vspace{2pt}\noindent\textbf{Experimental Setup.}\hspace{1ex}
We adopted a 7-class Kather-CRC-2016~\cite{kather2016multi} classification task (background filtered) under a strict few-shot setup. To synthesize relevant data, Gemini-2.5 Pro generated rich, descriptive prompts from representative images for each class. $\ours$ and all baselines used these prompts for augmentation. We report the mean F1-Weighted score over 5 random seeds on the real test set.

\vspace{2pt}\noindent\textbf{Quantitative Analysis.}\hspace{1ex}
The results in Figure~\ref{fig:tier4} illustrate the superior utility of data generated by $\ours$. Figure~\ref{fig:tier4} (Left) shows that `Real Data + $\ours$'' significantly outperforms the ``Real Data'' baseline across all K-shot settings, which confirms its augmentation effectiveness. Figure~\ref{fig:tier4} (Right) shows that among SOTA methods, only $\ours$ and Show-o2 deliver positive F1 gains at every K, and $\ours$ shows a clear lead, especially in the extreme few-shot, with larger gains than Show-o2 and PixCell. Other models yield negative gains, which indicates that their synthetic data degrades performance.

\begin{figure*}[tp]
    \centering
    \includegraphics[width=\linewidth]{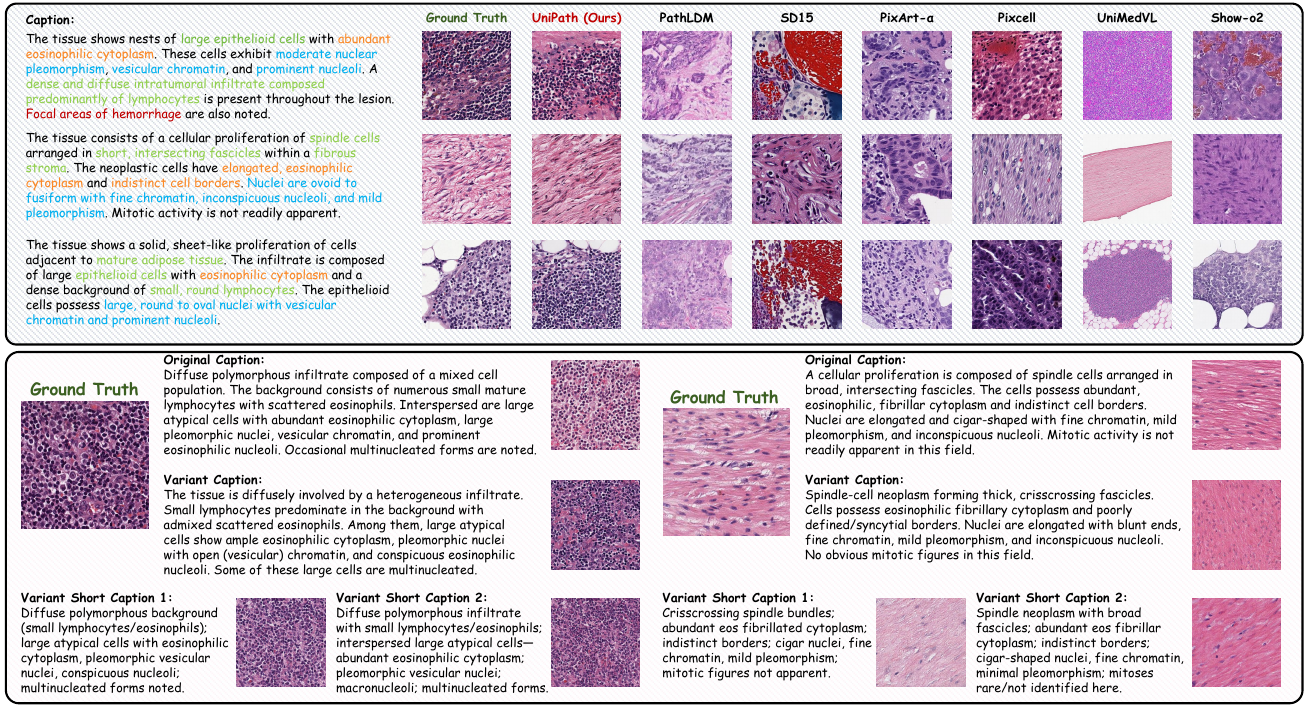}
    \caption{
        \textbf{Qualitative examples of $\ours$'s generation and controllability.} 
        \textbf{(Top)} Comparison of $\ours$ with SOTA baselines on complex prompts. Prompt text is color-coded by pathological concept: (\textcolor{MyGreen}{Tissue/Cell Type}, \textcolor{MyBlue}{Nuclear Features}, \textcolor{MyOrange}{Cytoplasm}, \textcolor{MyRed}{Hemorrhage}).
        \textbf{(Bottom)} Robustness of $\ours$ to synonymous prompts, validating the MSC's handling of terminological heterogeneity.
    }
    \label{fig:qualitative}
\end{figure*}

\subsection{Ablation Studies}
\label{sec:ablation}

We ablate our (i) data curation strategy and (ii) Multi-Stream Control (MSC) architecture. Figure~\ref{fig:ablation} plots all results on Visual Fidelity vs. Text-Image Alignment.

\vspace{2pt}\noindent\textbf{Ablation on Data Contribution.}\hspace{1ex} The \textcolor{myblue}{blue path} in Figure~\ref{fig:ablation} shows our data ablation. Starting from the TCGA (1.62M) baseline, adding the 1.03M HISTAI dataset significantly improves both metrics. The final 50K fine-tuning dataset pushes performance further to achieve the best overall results, proving the value of our two-stage data strategy.

\vspace{2pt}\noindent\textbf{Ablation on MSC Architecture.}\hspace{1ex}
The \textcolor{myorange}{orange} and \textcolor{myred}{red} paths in Figure~\ref{fig:ablation} ablate the MSC architecture. Starting from the poor-performing RTS-only baseline, adding the HLS stream primarily boosts alignment, while adding the PS stream improves both metrics. The poor alignment of the HLS+PS variant proves that all streams are indispensable. Our $\ours$ leverages all three streams to achieve the optimal balance, firmly occupying the ``Better'' region. We also provide inference-time ablations in \textbf{Appendix~\ref{sec:appendix_msc_ablation}}, analyzing the sensitivity of key MSC hyperparameters.

\section{Qualitative Analysis and Controllability}
\label{sec:qualitative}

We provide key qualitative visual evidence in Figure~\ref{fig:qualitative} to intuitively demonstrate $\ours$'s SOTA performance and the effectiveness of our MSC architecture in addressing core challenges, such as terminological heterogeneity. More diverse examples are provided in \textbf{Appendix~\ref{sec:appendix_qual}}.

\vspace{2pt}\noindent\textbf{SOTA Qualitative Comparison.}\hspace{1ex}
Figure~\ref{fig:qualitative} (top) compares $\ours$ against SOTA baselines on complex prompts. Baseline models often exhibit concept dropping, introduce artifacts, or render erroneous morphologies when faced with prompts containing multiple pathological features. In contrast, $\ours$ is the only model capable of accurately and high-fidely reproducing all specified pathological features. This qualitative result visually confirms our SOTA Visual Fidelity and highlights the CLIP-Score's limitations. While Show-o2 scored slightly higher, this figure proves $\ours$ has superior practical semantic alignment by rendering fine-grained concepts that the metric missed.

\vspace{2pt}\noindent\textbf{MSC: Robustness to Paraphrasing.}\hspace{1ex}
Figure~\ref{fig:qualitative} (bottom) visually validates our MSC's ability to resolve terminological heterogeneity. As shown in the samples, $\ours$ generates morphologically consistent images for ``Original'' and ``Variant'' captions despite different phrasing. This demonstrates the efficacy of $\ours$'s multi-stream design: the HLS stream distills heterogeneous prompts into the same diagnostic semantic tokens, maintaining stable semantic guidance and achieving robust control.
\section{Conclusion}
\label{sec:conclusion}

In this paper, we introduce $\ours$, a novel unified pathology model coupling MLLM understanding with a DiT generator. Our core contribution is the Multi-Stream Control architecture, which uses a High-Level Semantic stream and a Prototype stream to tackle terminological heterogeneity and component-level morphological control simultaneously. To support and validate $\ours$, we curated a large-scale corpus, a 68K high-fidelity subset, and established a four-tier evaluation protocol. Extensive experiments show $\ours$ demonstrates leading performance in both automated metrics and human expert evaluations. 

\vspace{2pt}\noindent\textbf{Broader Impacts and Future Work.}\hspace{1ex}
As a controllable, unified pathology foundational model, $\ours$ offers significant potential in data augmentation by generating high-fidelity, customized synthetic images; in research by enabling the systematic exploration of morphological features; and in education as an interactive training tool. For future work, we plan to extend $\ours$ to support higher-resolution image generation and explore its capabilities in pathological image editing. 
We provide further discussions of the limitations and ethical issues in the \textbf{Appendix~\ref{sec:appendix_limitations}~\&~\ref{sec:appendix_ethics}}.

\clearpage

\bibliographystyle{plainnat}
\bibliography{main}

\clearpage

\beginappendix
\appendix
\setcounter{figure}{0}
\setcounter{table}{0}
\renewcommand{\thefigure}{S\arabic{figure}}
\renewcommand{\thetable}{S\arabic{table}}

\renewcommand{\contentsname}{Appendix Contents}

\tableofcontents

\addtocontents{toc}{\protect\setcounter{tocdepth}{2}}

\addtocontents{toc}{\protect\setlength{\protect\cftsecnumwidth}{1.8em}}

\vspace{2em} 

\section{Limitations}
\label{sec:appendix_limitations}

\vspace{2pt}\noindent\textbf{Text–Image Alignment Performance.}\hspace{1ex}
As shown in Table~\ref{tab:tier1&2}, $\ours$ is SOTA on every downstream metric, on Real2Gen retrieval, and in the MLLM‑as‑Judge comparison, yet trails Show‑o2 on the single CONCH CLIP‑Score (0.348 vs. 0.357). We argue that this gap reflects a bias in the evaluator–model paradigm, rather than a true semantic alignment weakness.

\begin{itemize}[leftmargin=*]
\item \textbf{Show‑o2’s semantic tokens are contrastive‑distilled}.  
      During training, Show‑o2 loads SigLIP weights into its Semantic Layers and minimizes a distillation loss, forcing those tokens to mimic SigLIP patch features.  
      Although generation itself uses flow matching, the resulting high‑level tokens remain ``SigLIP‑style’’ at inference.
      
\item \textbf{The evaluator is also contrastive.}  
      CONCH~\cite{conch} is a CLIP‑family model whose similarity metric is the same cosine space SigLIP is trained in.  
      A model whose internal tokens are pre‑aligned to this space naturally receives a higher CLIP score.

\item \textbf{$\ours$ uses MLLM‑derived pathology semantics without SigLIP distillation.}  
      Our HLS stream extracts diagnosis tokens from a frozen Patho‑R1 MLLM.  
      Patho‑R1’s vision encoder is CLIP‑based, but no part of the generator is forced to match CLIP/SigLIP features; the DiT learns purely via flow‑matching reconstruction.  
      Hence, its latent space is optimised for morphological fidelity, not for cosine similarity with CLIP evaluators.

\item \textbf{Cross‑metric consistency.}  
      $\ours$ leads on Real2Gen retrieval (closer in feature space to real WSIs), on MLLM‑as‑Judge human‑preference scoring, and delivers the largest Tier 4 F1 gains.  
      These orthogonal results confirm that the small CLIP‑Score gap is an artefact of evaluator homology, not of inferior text–image alignment.
\end{itemize}

In summary, the lower CONCH CLIP‑Score stems from Show‑o2’s SigLIP‑distilled tokens matching the evaluator’s contrastive space, whereas $\ours$ prioritises pathology‑specific morphology and semantics, which better serve real diagnostic tasks.

\vspace{2pt}\noindent\textbf{Dependency on Prototype Bank.}\hspace{1ex}
One of $\ours$'s strengths comes from the component-level control provided by the Prototype Stream (PS). This advantage, however, is highly dependent on the quality and coverage of our 8K instance prototype bank. If an extremely rare morphological component is not well-represented in our 8K bank, the PS stream cannot provide precise control for that concept.

\section{Ethical Statement}
\label{sec:appendix_ethics}

This research strictly adheres to the relevant ethical guidelines for medical AI research.

\vspace{2pt}\noindent\textbf{Data Usage and Patient Privacy.}\hspace{1ex}
All data used in this study (TCGA and HISTAI) are publicly available datasets intended for research. All data were fully anonymized and de-identified by the original providers prior to release and contain no Protected Health Information (PHI). Our usage strictly complies with the Data Use Agreements (DUAs) for both TCGA and HISTAI.

\vspace{2pt}\noindent\textbf{Potential for Misuse and Mitigation.}\hspace{1ex}
We acknowledge that high-fidelity pathological image generation (\ie, ``medical deepfakes'') carries a potential risk of misuse, such as attempting to interfere with clinical diagnostic workflows in extreme cases. We emphasize that $\ours$ is currently intended for research purposes only, with the design goals of (i) advancing controllable generation in pathology, (ii) providing controllable data augmentation for computational pathology, and (iii) serving as an educational tool. This model \textbf{must not} be used for any direct clinical diagnosis.

\vspace{2pt}\noindent\textbf{Algorithmic Bias.}\hspace{1ex}
Our model's performance relies on the quality and distribution of our training data. Despite our efforts to add diversity (1.03M HISTAI) and balance our 68K subset (via K-means and elite sampling), our training data may still contain undiscovered biases (\eg, in demographic representation across race, age, or sex). The model may learn and amplify these biases. Future work is required to specifically quantify and mitigate such biases.

\section{Future Work}
\label{sec:appendix_feature_work}

While $\ours$ marks significant progress in unifying pathology understanding and controllable synthesis, several key directions remain for future exploration.

\vspace{2pt}\noindent\textbf{Support for Higher Resolution and Broader Histological Context.}\hspace{1ex}
The current $\ours$ model primarily operates on $384 \times 384$ pixel patches. While sufficient for capturing cell-level morphological features, this limits the model's ability to understand and generate larger-scale architectural patterns, such as complex glandular structures or tumor-stroma interactions. Future work should explore extending $\ours$ to higher resolutions or integrating a larger field of view, enabling the generation of images that are more histologically context-aware.

\vspace{2pt}\noindent\textbf{Controllable Pathological Image Editing.}\hspace{1ex}
$\ours$ currently focuses on image generation from text prompts. A high-impact extension is to enable fine-grained editing of existing real pathology images. This can be framed as a ``counterfactual synthesis'' task — such as ``adding moderate nuclear atypia'' to a benign tissue image or ``removing the specified inflammatory infiltrate.'' The MSC architecture of $\ours$ provides an ideal framework for this: the HLS stream could parse the editing instruction (\eg, ``increase mitotic figures''), while the PS stream could retrieve and inject the corresponding morphological prototypes to achieve the precise, localized modification.

\vspace{2pt}\noindent\textbf{Scaling the Prototype Bank.}\hspace{1ex}
Our prototype-based control mechanism opens a promising avenue for future enhancement. While the curated 8K bank establishes the efficacy of this approach, we can further enhance the model's generative ``vocabulary'' by scaling this bank. Future work could explore using active learning or self-supervised methods to automatically mine and cluster novel, informative prototypes from large-scale, unlabeled datasets. This expansion would enable $\ours$ to synthesize an even greater diversity of morphological features with high precision, particularly for rare, long-tail pathological phenomena.

\section{Detailed Implementation Details}
\label{sec:appendix_implementation}

\subsection{Model Architecture Specifications}
\label{sec:appendix_arch}

\vspace{2pt}\noindent\textbf{Generation Backbone (DiT) and Conditioning.}\hspace{1ex}
Our generation backbone is a 0.6B-parameter DiT (Diffusion Transformer) designed in the PixArt~\cite{chen2023pixart} style. It comprises 28 Transformer layers, 16 attention heads, and a hidden dimension $d=1152$. Our model employs a hybrid conditioning mechanism. The fused conditional vector $C_{comp}$ is injected into every DiT layer via traditional cross-attention. In contrast, the timestep is handled separately, injected via the AdaLayerNorm-Single (AdaLN-S) mechanism to perform conditional normalization.

\vspace{2pt}\noindent\textbf{Understanding Backbone and VAE.}\hspace{1ex}
Our backbone is based on the Patho-R1 (7B)~\cite{pathor1} model, which is post-trained on pathology domain data from Qwen2.5-VL 7B~\cite{bai2025qwen2}. The backbone remains fully frozen throughout all training stages. The VAE employed is the Stable Diffusion 3~\cite{sd3} VAE, featuring an 8x downsampling factor.

\vspace{2pt}\noindent\textbf{MSC Module Implementation.}\hspace{1ex}
In the Multi-Stream Control (MSC) module, the projection layers for the three streams (HLS, RTS, and PS), such as $\mathrm{MLP}_{\text{DST}}$, are implemented as separate 2-layer Feed-Forward Networks (FFNs) with unshared weights. Each of these MLPs follows the same architecture: a linear projection from the input dimension to the hidden size, followed by a GELU activation, and a second linear projection from the hidden size back to the hidden size. The hidden size is 1152 (matching the DiT's hidden dimension).
For the Prototype Stream (PS) retrieval, we provide the specific hyperparameters used for the hybrid strategy. The total number of prototypes is $K_m=16$. For the Global Semantic Retrieval ($U_g$, Eq. 4), we set the retrieval tops $k_t=4$ (Text) and $k_v=4$ (Vision). For the Local Fine-grained Retrieval ($U_l$), we parse the four rarest keywords from the prompt and randomly sample 2 prototypes for each term, resulting in 8 local prototypes. The final set $\hat{U}$ is the union of $U_g$ and $U_l$, clipped to 16 (Eq. 5).

\vspace{2pt}\noindent\textbf{Flow Matching Training and Inference.}\hspace{1ex}
During the training stage, we adopt a Rectified Flow strategy. Specifically, we first sample $u \sim \mathcal{U}(0, 1)$, map it to timesteps and sigmas, and construct the noise-interpolated $z_t$ accordingly. The model $v_\theta$ is trained to predict the target velocity $v_t = z_0 - z_1$.
During the inference stage, we use the Euler solver with 30 function evaluations. We employ Classifier-Free Guidance with a guidance\_scale of 3.0.

\subsection{Training Hyperparameters}
\label{sec:appendix_hyperparams}

\vspace{2pt}\noindent\textbf{General Setup.}\hspace{1ex}
Across both training stages, we used the AdamW optimizer with default betas ($\beta_1=0.9, \beta_2=0.999$), an epsilon of $1e^{-8}$, and a weight decay of 0.01. All training was conducted using mixed precision. All experiments were conducted on 16 NVIDIA H100 GPUs. All input images were processed to a resolution of $384 \times 384$.

\vspace{2pt}\noindent\textbf{Stage 1: Semantic Alignment (Pre-training).}\hspace{1ex}
The model was pre-trained on 2.58M text-image pairs (excluding the 68K subset) for 10,000 steps using the Flow Matching (MSE) loss. We used a global batch size of 512. The learning rate was linearly warmed up for the first 2\% of steps (200 steps) to a peak of 1e-4. It was then decayed using a cosine scheduler with a minimum learning rate of $1e^{-5}$.

\vspace{2pt}\noindent\textbf{Stage 2: High-Quality Fine-tuning.}\hspace{1ex}
The model was subsequently fine-tuned on the 50K high-quality subset for 500 steps. We used a global batch size of 512 and a fixed learning rate of 2e-5 for this entire stage.

\section{Dataset Detailed Analysis}
\label{sec:appendix_dataset}

\subsection{Statistics of the 1.03M Corpus}
\label{sec:appendix_dataset_1M}

\begin{figure*}[htp]
  \centering
  \includegraphics[width=1.0\textwidth]{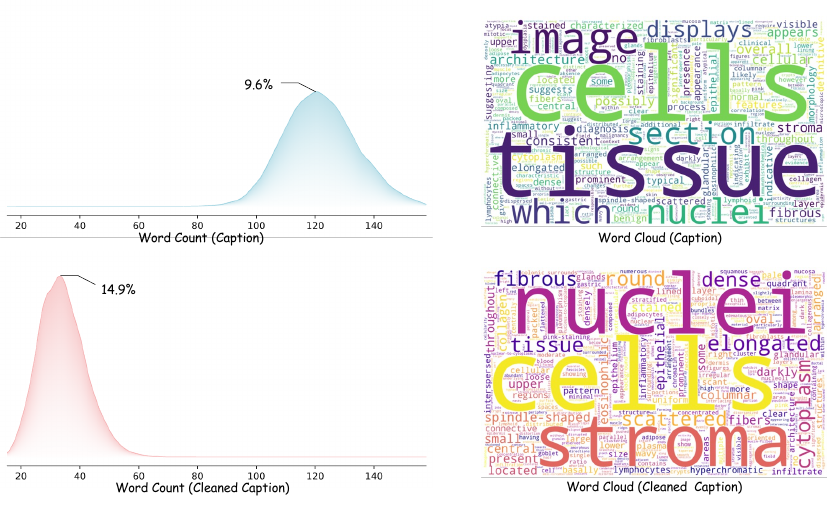}
  \caption{Visualization of the caption-length distribution generated by PathGen-LLaVa (top left) and its corresponding word-frequency cloud (top right), as well as the caption-length distribution after summarization by Qwen3-8B (bottom left) and the associated word-frequency cloud (bottom right).}
  \label{fig:appendix_word_analysis1}
\end{figure*}

\vspace{2pt}\noindent\textbf{Word Count.}\hspace{1ex}
We analyzed the caption-length distributions of the 1.03M corpus before and after summarization using Qwen3-8B, as shown on the left side of Figure~\ref{fig:appendix_word_analysis1}. The distributions exhibit unimodal and symmetric curves. In the original captions, the most frequent length is around 120 words, accounting for 9.6\%. After refinement, the peak shifts to approximately 35 words, accounting for 14.9\%, whereas captions longer than 60 words are virtually absent.

\vspace{2pt}\noindent\textbf{Word Frequency.}\hspace{1ex}
We analyzed the word-frequency profiles of captions before and after cleaning the 1.03M corpus, which are presented as word clouds in the right panel of Figure~\ref{fig:appendix_word_analysis1}. The word clouds indicate that the captions emphasize microscopic morphological features such as ``cells,'' ``nuclei,'' and ``stroma,'' as well as diagnostic descriptors including ``inflammatory'' and ``stained.'' A comparison of the two word clouds shows an increased prevalence of morphology-related terms and a marked reduction in non-informative tokens such as ``which'' and ``image.''

\subsection{Analysis of the 68K Refined Subset}
\label{sec:appendix_dataset_68K}

\begin{figure*}[htp]
  \centering
  \includegraphics[width=1.0\textwidth]{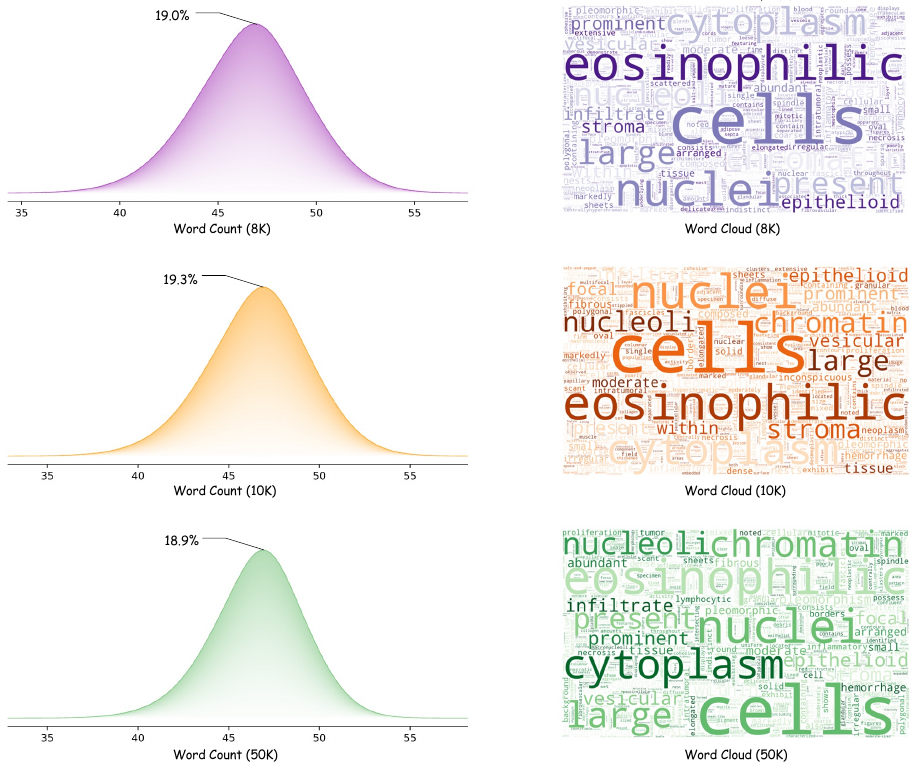}
  \caption{Visualization of the caption-length distributions and word-frequency clouds for the 8K, 10K, and 50K refined subsets (8K: top row; 10K: middle row; 50K: bottom row), with length distributions shown on the left and word-frequency clouds on the right.}
  \label{fig:appendix_word_analysis2}
\end{figure*}

\vspace{2pt}\noindent\textbf{Word Count.}\hspace{1ex}
We analyzed the caption lengths in the 68K Refined Subset, as shown on the left of the Figure~\ref{fig:appendix_word_analysis2}. The distribution is symmetric and unimodal, with the most frequent length around 47 words, accounting for approximately 19\%. Compared with the 1.03M Corpus, captions in the 68K subset are more extended, rarely shorter than 35 or longer than 55 words, due to the prompt-imposed 30–60-word constraint. This moderate length reduces redundancy while ensuring sufficient content to accurately describe the images, enabling the model to learn a broader range of knowledge.

\vspace{2pt}\noindent\textbf{Word Frequency.}\hspace{1ex}
We also analyzed the word-frequency distribution of captions in the 68K Refined Subset, visualized as word clouds on the right side of Figure~\ref{fig:appendix_word_analysis2}. The three subsets (8K, 10K, and 50K) exhibit a highly coherent vocabulary profile dominated by morphological, nuclear, cytoplasmic, stromal, and diagnostic descriptors. Unlike the 1M Corpus, where terms such as ``nuclei,'' ``cells,'' and ``stroma'' overwhelmingly dominate, the 68K Refined Subset exhibits a more balanced distribution of key pathological concepts. The captions in this refined subset are more detailed and make use of a richer and more uniformly distributed set of domain-specific terms, thereby providing higher-quality supervision that is advantageous for model evaluation and fine-tuning.

\begin{table*}[!t]
  \caption{Overall results of models on the PathMMU \textbf{test set}. The best-performing MLLM in each subset for general and medical/pathology MLLM is \textbf{in-bold}, and the second-best performing MLLM is {\underline{underlined}}. An asterisk ($*$) indicates results copied from the original source for the non–open-sourced model.}
  \vspace{-5pt}
  \centering
  \resizebox{\linewidth}{!}{
    \begin{tabular}{@{}lcccccccccccccc@{}}
      \toprule
      \textbf{} & \multirow{3}{*}{\textbf{\makecell{Unif.\\Model}}} & \multicolumn{2}{c}{\textbf{Test Overall}} & \multicolumn{2}{c}{\textbf{PubMed}} & \multicolumn{2}{c}{\textbf{SocialPath}} & \multicolumn{2}{c}{\textbf{EduContent}} & \multicolumn{2}{c}{\textbf{Atlas}} & \multicolumn{2}{c}{\textbf{PathCLS}} \\ 
      & & Tiny  & ALL  & Tiny  & ALL & Tiny  & All & Tiny  & All  & Tiny  & ALL  & Tiny  & ALL\\
      & & (1131)  & (9483)  & (281)  & (3068) & (210)  & (1661) & (255)  & (1938)  & (208)  & (1007) & (177)  & (1809)\\
      \midrule
    \rowcolor{dino}
    \multicolumn{1}{l}{\textcolor{gray}{Expert Performance}} &
    \textcolor{gray}{-} & \textcolor{gray}{71.8} & \textcolor{gray}{-} & \textcolor{gray}{72.9} &
    \textcolor{gray}{-} & \textcolor{gray}{71.5} & \textcolor{gray}{-} & \textcolor{gray}{69.0} &
    \textcolor{gray}{-} & \textcolor{gray}{68.3} & \textcolor{gray}{-} & \textcolor{gray}{78.9} &
    \textcolor{gray}{-} \\

      \midrule
      \multicolumn{14}{c}{\textbf{\textit{General Multimodal Large Language Models}}} \\ \midrule
      LLaVA-1.5-13B   & \textcolor{red!80!black}{\ding{56}} & 39.1 & 37.7 & 45.2 & 41.4 & 39.6 & 40.1 & 35.0 & 39.9 & 46.5 & 43.4 & 25.9 & 23.7  \\
      LLaVA-OneVision-7B & \textcolor{red!80!black}{\ding{56}}& 36.8& 34.0& 38.2 & 36.9 & 34.9 & 37.7 & 46.4 & 37.4 & 39.6 & 38.8 & 20.0 & 19.5  \\
      Qwen3-VL-8B-Instruct  &\textcolor{red!80!black}{\ding{56}} & 54.9 & 51.5 & 59.1 & 54.4 & 63.3 & 56.3 & 56.1 & 53.7 & 52.4 & 55.1 & 39.5 & 37.9  \\
      Qwen3-VL-30B-A3B-Instruct & \textcolor{red!80!black}{\ding{56}}& 57.6 & 54.5 & 64.4 & 57.2 & 67.6 & 65.8 & 58.0 & 56.1 & 56.3 & 58.3 & 35.6 & 35.7  \\
      BLIP3o-8B & \textcolor{green!60!black}{\ding{52}}&  40.0& 37.6& 40.5 & 40.6 & 42.6 & 39.2 & 47.5 & 42.3 & 40.7 & 38.7&24.7 &25.3   \\
      Show-o2-7B &\textcolor{green!60!black}{\ding{52}} & 40.8 & 40.4 & 43.7 & 48.8& 43.8 & 40.4 & 42.4 & 41.9 & 42.8 & 40.4 &28.2 & 24.4  \\
      BAGEL-14B &\textcolor{green!60!black}{\ding{52}} & 57.3 & 52.3 & 63.0 & 56.2 & 61.4 & 59.6 & 60.8 & 54.1 & 64.4 & 61.3& 29.9&  31.9 \\
      GPT-4V-1106  & - & 53.8 & 49.9& 59.7 & 53.9 & 58.5 & 53.4 & 60.6 & 53.9 & 47.6 & 52.6 & 36.6 & 33.9\\
      Gemini-2.5 Pro    & -  & \textbf{69.0} & \textbf{68.0} & \textbf{74.5} &\textbf{71.8} &\textbf{72.3} &\textbf{68.9}& \textbf{72.4} &69.8& \underline{66.8} &69.5 &53.9 &\underline{58.1}   \\
      \midrule
      \multicolumn{14}{c}{\textbf{\textit{Medical / Pathology-specific Multimodal Large Language Models}}} \\ \midrule
      LLaVA-Med & \textcolor{red!80!black}{\ding{56}} & 25.5 &26.8 & 29.2 & 28.5 & 29.8 & 28.2 & 23.3 & 27.4 & 21.7 & 30.4 & 22.0 & 20.1\\
      Quilt-LLaVA    & \textcolor{red!80!black}{\ding{56}} & 45.4& 41.2& 46.8 & 41.9 & 46.3 & 45.9 & 51.0 & 45.0 & 46.5 & 43.7 & 32.8 & 30.1\\
      PathGen-LLaVA & \textcolor{red!80!black}{\ding{56}}& 59.8 &58.4 & 59.3 & 59.9 & 60.2 & 58.7 & 60.1 & 60.5 & 64.3 & 65.1 & \underline{54.4} & 49.7\\
      CPath-Omni$^{*}$ &\textcolor{red!80!black}{\ding{56}} & - & - & \underline{74.0} & \underline{69.9} & - & - & 69.8 & \underline{70.6} & 65.9 & \underline{70.6} & \textbf{75.7} & \textbf{79.0}\\
      UniMedVL &\textcolor{green!60!black}{\ding{52}} &54.5 &50.6 & 58.0 & 54.9 & 57.1 & 56.7 & 55.7 & 56.0 & 66.3 & 59.4 & 30.5&  26.9 \\
      \rowcolor{aliceblue} \textbf{$\ours$ (Ours)} &\textcolor{green!60!black}{\ding{52}} & \underline{68.3} & \underline{65.7} & 72.9 & 66.4 & \underline{67.9} & \underline{68.4} & \underline{70.1} & \textbf{73.9} & \textbf{79.2} & \textbf{77.7} & 46.1 & 46.6\\
      \bottomrule
    \end{tabular}}
  \label{tab:overall_results_pathmmu}
  \vspace{-2pt}
\end{table*}

\subsection{Spot-Check Validation of the 10K Test Set}
\label{sec:appendix_qc}
To definitively validate the reliability of the ``Gemini-2.5 Pro generation then GPT-5 review'' automated pipeline, we additionally invited a domain-expert pathologist to conduct an independent spot-check quality control (QC) on 500 random samples from our 10K high-quality test set.

The reviewer's task was to assign each image-text pair to one of three categories.
\textbf{3: Excellent} was defined as: The description is accurate, comprehensive, and professional, perfectly corresponding to all key pathological features in the image (Gold Standard).
\textbf{2: Acceptable} was defined as: The description captures the main diagnostic features without factual errors, but may contain minor deficiencies, such as omitting a secondary feature, slight imprecision in non-critical terminology, or a minor deviation in descriptive focus (Still Usable for Evaluation).
\textbf{1: Unusable} was defined as: The description contains severe factual errors, rendering it unsuitable as an evaluation benchmark, such as hallucinating key features not present in the image, misidentifying the primary cell type, or completely omitting the main diagnostic point of the image (Failure).

Upon reviewing the 500 random samples, the pathologist's evaluation was as follows: 43.4\% of the image-text pairs were rated as 3: Excellent; 50.2\% were rated as 2: Acceptable; and 6.4\% were rated as 1: Unusable. This results in an Overall Usability Rate (\ie, Excellent + Acceptable) of 93.6\%. This extremely low ``Unusable'' rate (6.4\%) strongly confirms the SOTA reliability of the automated data annotation pipeline we employed.

\subsection{1231-term Pathology Vocabulary}
\label{sec:appendix_vocabulary}
Our 1231-term pathology vocabulary, which was used to build the inverted index $\mathcal{I}$, is provided as a separate file (``vocabulary.txt'') in the supplementary material bundle.

\subsection{Analysis of Data Leakage Risks}
\label{sec:appendix_risks}
To ensure the integrity of our evaluation splits and proactively address potential concerns regarding overlap between the 8K Prototype Bank and the 10K Test Set, we conducted a rigorous validation.
We computed the exhaustive pairwise visual feature similarity ($8,000 \times 10,000 = 80,000,000$ comparisons) between all images in the bank and all images in the test set. For this check, we used the UNI2-h extractor~\cite{uni}, which is the same high-performance backbone we employed for dataset-wide de-duplication in Section~\ref{sec:large_scale_data}.
The statistics confirmed that the sets are strictly disjoint: the average similarity was 0.1358 (standard deviation = 0.0686). Critically, the maximum similarity observed across all 80 million pairs was 0.9416.
This maximum value (0.9416) is below our defined de-duplication threshold of 0.95 (as detailed in Section~\ref{sec:large_scale_data}). This result strongly confirms that our prototype bank and test set are strictly disjoint. It therefore eliminates any risk of data leakage via the Prototype Stream (PS) retrieval, validating the integrity of our Tier 2 (Alignment) and Tier 3 (Control) evaluations.

\section{Supplementary Quantitative Results}
\label{sec:appendix_quant}

\subsection{Full Understanding Capability}
\label{sec:appendix_tier0}

To validate the diagnostic understanding capability of $\ours$, which is critical for our Multi-Stream Control (MSC) module, we evaluated its performance on the comprehensive PathMMU benchmark~\cite{sun2024pathmmu}. The full results are presented in Table~\ref{tab:overall_results_pathmmu}.
As shown in the table, $\ours$ achieves an overall score of 65.7 on the full test set. This performance establishes $\ours$ as the SOTA among all evaluated open-source models, substantially outperforming other leading open-source pathology MLLMs, including PathGen-LLaVA (58.4) and the unified model UniMedVL (50.6).
Furthermore, $\ours$ achieves the top score across all models (including closed-source systems) on the EduContent (73.9) and Atlas (77.7) sub-tasks. Its overall score also closely approaches that of top-tier closed-source models, such as Gemini-2.5 Pro (68.0). This strong understanding performance confirms that our frozen MLLM backbone provides the robust, phrasing-invariant semantics necessary to steer controllable generation.

\subsection{Additional Fidelity \& Alignment Results}
\label{sec:appendix_more_metrics}

{\renewcommand{\arraystretch}{1.1}

\begin{table*}[tp]
\centering
\caption{Quantitative comparison of Visual Fidelity and Text--Image Alignment with merged T2I/I2I. FID/KID uses the Virchow2 extractor; Similarity and retrieval metrics use MUSK. \textsuperscript{\textcolor{orange}{\scalebox{1.2}{\ding{72}}}} marks models fully fine-tuned on our large dataset. The best is \textbf{bold}, the second best is \underline{underlined}.}
\label{tab:tier1&2_supp}
\resizebox{\linewidth}{!}{
\begin{tabular}{lccc|ccccc}
\toprule
\multirow{2}{*}{\textbf{}} & \multirow{2}{*}{\textbf{\makecell{Unif.\\Model}}} & \multicolumn{2}{c|}{\textbf{Visual Fidelity}~~$\DOWN$} & \multicolumn{5}{c}{\textbf{Text--Image Alignment}~~$\UP$} \\
\cline{3-9}
& \rule{0pt}{3ex} & \textbf{FID} & \textbf{KID} & \textbf{Sim.} & \textbf{Recall@10} & \textbf{Recall@50} & \textbf{mAP@10} & \textbf{mAP@50} \\
\hline
\rowcolor{dino}
\multicolumn{1}{l}{\textcolor{gray}{Real Data}} &
\textcolor{gray}{-} & \textcolor{gray}{-} & \textcolor{gray}{-} &
\textcolor{gray}{0.557} &
\textcolor{gray}{13.40/--} & \textcolor{gray}{33.50/--} &
\textcolor{gray}{5.28/--} & \textcolor{gray}{6.17/--} \\
\hline
\multicolumn{9}{c}{\textbf{\textit{General Text to Image Generation Models}}} \\
\hline
SD1.5\textsuperscript{\textcolor{orange}{\scalebox{1.2}{\ding{72}}}}~\cite{ldm}   & \textcolor{red!80!black}{\ding{56}} & 1804.69 & 0.519 & 0.483 & 0.66/0.60 & 1.91/1.94 & 0.80/0.80 & 0.85/0.91 \\
SDXL\textsuperscript{\textcolor{orange}{\scalebox{1.2}{\ding{72}}}}~\cite{podell2023sdxl} & \textcolor{red!80!black}{\ding{56}} & 2570.19 & 0.602 & 0.445 & 0.34/0.22 & 1.20/0.80 & 0.60/0.38 & 0.77/0.49 \\
Pixart-$\alpha$\textsuperscript{\textcolor{orange}{\scalebox{1.2}{\ding{72}}}}~\cite{chen2023pixart} & \textcolor{red!80!black}{\ding{56}} & 2574.85 & 0.685 & 0.482 & 1.14/0.60 & 4.54/2.65 & 1.58/0.72 & 1.74/0.93 \\
BLIP3o\textsuperscript{\textcolor{orange}{\scalebox{1.2}{\ding{72}}}}~\cite{blip3o} & \textcolor{green!60!black}{\ding{52}} & 2008.75 & 0.550 & 0.455 & 1.55/1.50 & 5.87/5.80 & 2.39/1.93 & 2.58/2.26 \\
Show-o2\textsuperscript{\textcolor{orange}{\scalebox{1.2}{\ding{72}}}}~\cite{showo2} & \textcolor{green!60!black}{\ding{52}} & 1398.52 & 0.415 & \textbf{0.545} & \textbf{8.41}/\underline{2.71} & \textbf{22.77}/\underline{9.74} & \textbf{10.93}/\underline{3.22} & \textbf{9.42}/\underline{3.11} \\
\hline
\multicolumn{9}{c}{\textbf{\textit{Pathological / Medical Text to Image Generation Models}}} \\
\hline
Pixcell~\cite{yellapragada2025pixcell}  & \textcolor{red!80!black}{\ding{56}} & \underline{929.30} & \underline{0.259} & 0.524 & - & - & - & - \\
PathLDM~\cite{yellapragada2024pathldm} & \textcolor{red!80!black}{\ding{56}} & 1126.36 & 0.376 & 0.483 & 0.15/0.17 & 0.73/0.70 & 0.18/0.25 & 0.30/0.36 \\
UniMedVL~\cite{ning2025unimedvl}        & \textcolor{green!60!black}{\ding{52}} & 1435.07 & 0.363 & 0.520 & 3.82/2.23 & 11.52/7.18 & 5.31/2.59 & 5.14/2.74 \\
\rowcolor{aliceblue}\textbf{$\ours$ (Ours)} & \textcolor{green!60!black}{\ding{52}} & \textbf{484.38} & \textbf{0.192} & \underline{0.538} & \underline{7.55}/\textbf{4.25} & \underline{21.61}/\textbf{13.72} & \underline{10.22}/\textbf{6.38} & \underline{8.93}/\textbf{6.07} \\
\bottomrule
\end{tabular}
}
\end{table*}}

In our main evaluation (Section~\ref{sec:expermiments}), the text-image alignment metrics and Patho-FID/KID metrics were based on the CONCH and UNI2-h backbones, respectively. 
To further validate the robustness and generality of our findings, we conducted an additional evaluation using two entirely independent, external backbones not used anywhere in our model pipeline: Virchow2~\cite{virchow} for Visual Fidelity and MUSK~\cite{musk} for Text-Image Alignment. 
This analysis confirms that our model's superior performance is a genuine advantage and not an artifact of a specific evaluator. The results are shown in Table~\ref{tab:tier1_2_supp}.

\vspace{2pt}\noindent\textbf{Visual Fidelity (Virchow2 Backbone).}\hspace{1ex}
The results using the Virchow2 feature extractor strongly reinforce our main findings. $\ours$ achieves a FID of 484.38 and a KID of 0.192. This performance is not just SOTA, but represents a massive improvement over the next-best model, Pixcell (FID: 929.30, KID: 0.259). This confirms that the superior visual fidelity of $\ours$ is a genuine model advantage, not an artifact of the UNI2-h evaluator.

\begin{figure}[tp]
  \centering
  \includegraphics[width=0.4\linewidth]{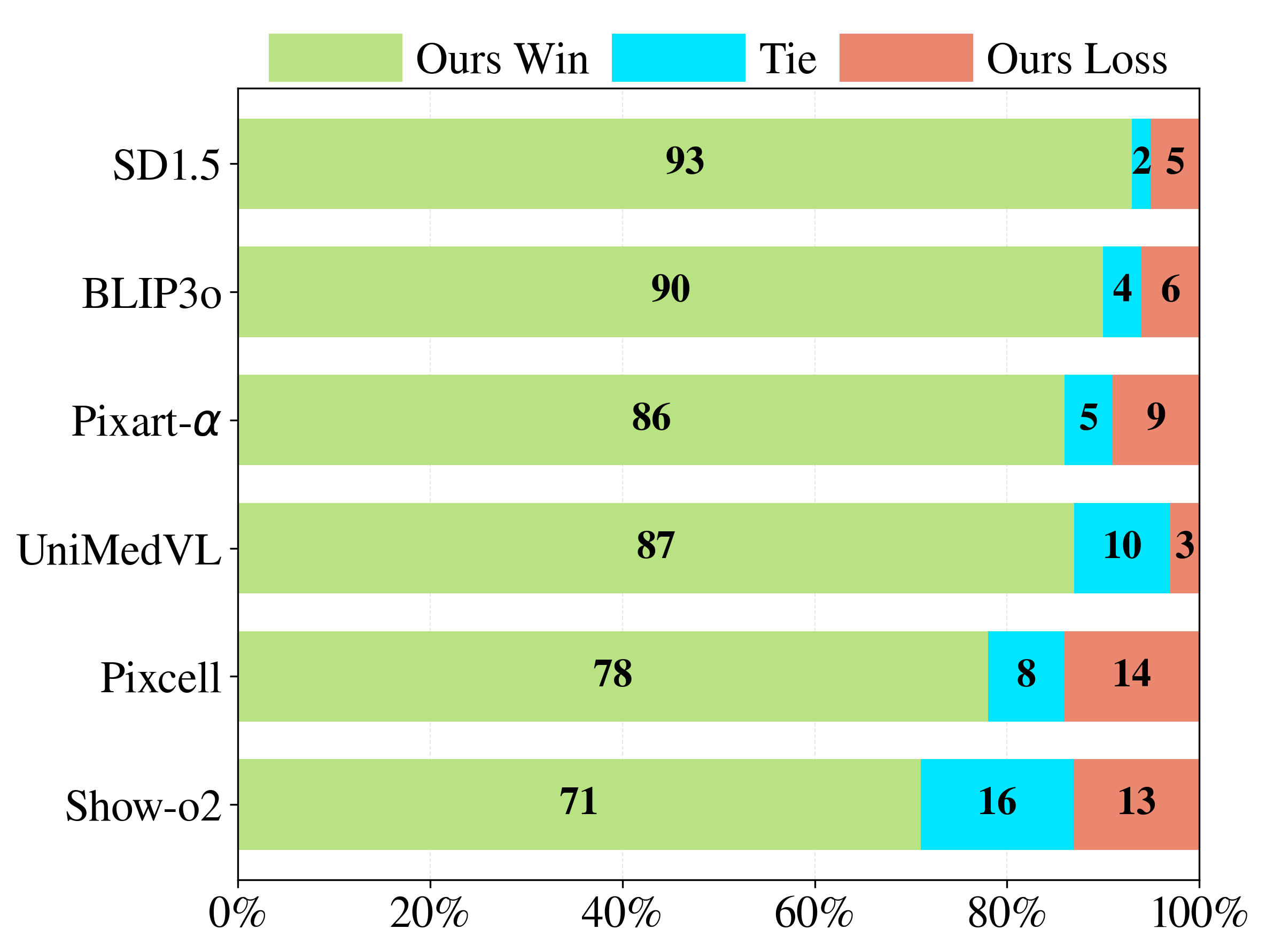}
  \caption{Gemini 2.5 Pro as Judge.}
  \label{fig:appendix_gemini_judge}
\end{figure}

\vspace{2pt}\noindent\textbf{Text-Image Alignment (MUSK Backbone).}\hspace{1ex}
The alignment metrics using the MUSK backbone provide a crucial, unbiased perspective.
\begin{itemize}[leftmargin=1.5em]
    \item \textbf{CLIP-Score \& T2I Retrieval:} Using MUSK, \textbf{Show-o2} achieves the highest CLIP-Score (0.545) and the best Report-Gen (T2I) retrieval metrics. This result is consistent with our main paper's findings (Table 1) and supports our hypothesis that Show-o2's architecture is well-optimized for general-purpose T2I alignment.
    \item \textbf{I2I Retrieval:} Critically, on the Real-Gen (I2I) retrieval task—which measures how close the generated images are to real images in this new feature space—$\ours$ achieves dominant SOTA performance across all four metrics (Recall@10/50, mAP@10/50). For instance, our mAP@10 (6.38) is nearly double that of the second-best Show-o2 (3.22).
\end{itemize}

\vspace{2pt}\noindent\textbf{Conclusion.}\hspace{1ex}
These results, obtained from fully independent feature extractors, strongly validate our conclusions. Our model's superior visual fidelity and its ability to generate images that are most faithful to the real pathology manifold are thus demonstrated as robust and general findings, independent of the specific evaluator used.

\subsection{Detailed Few-Shot Classification Results}
\label{sec:appendix_tier4_table}
{\renewcommand{\arraystretch}{1.2}

\begin{table*}[tp]
\centering

\caption{Few-shot downstream performance (Weighted F1) across different shots $K$. Values are Mean$\pm$Std (\%); ``$\Delta$'' \emph{columns} report absolute change vs.\ original data in percentage points. Best is \textbf{bold}, second best is \underline{underlined}.}
\label{tab:fewshot_wf1}
\resizebox{\linewidth}{!}{
\begin{tabular}{lcccccccccc}
\toprule
\multirow{2}{*}{\textbf{Model}} & \multicolumn{2}{c}{\textbf{K = 2}} & \multicolumn{2}{c}{\textbf{K = 4}} & \multicolumn{2}{c}{\textbf{K = 8}} & \multicolumn{2}{c}{\textbf{K = 16}} & \multicolumn{2}{c}{\textbf{K = 32}} \\
& Wgt. F1 & $\Delta$ & Wgt. F1 & $\Delta$ & Wgt. F1 & $\Delta$ & Wgt. F1 & $\Delta$ & Wgt. F1 & $\Delta$ \\
\hline
\multicolumn{11}{c}{\textbf{\textit{Baselines (Only Real Data)}}} \\
\hline
Original Data            & 67.34$_{\pm 2.37}$ & {\small --}   & 76.42$_{\pm 3.31}$ & {\small --}   & 81.43$_{\pm 1.88}$ & {\small --}   & 83.85$_{\pm 1.48}$ & {\small --}   & 86.88$_{\pm 0.88}$ & {\small --}   \\
\rowcolor{aliceblue} \textbf{$\ours$ (Ours)}        & 50.31$_{\pm 8.92}$ & {\small --}   & 69.33$_{\pm 2.81}$ & {\small --}   & 76.89$_{\pm 2.35}$ & {\small --}   & 78.30$_{\pm 0.99}$ & {\small --}   & 81.05$_{\pm 0.77}$ & {\small --}   \\
\hline
\multicolumn{11}{c}{\textbf{\textit{Data Augmented Comparisons (Real Data with Generated Data)}}} \\
\hline
SD1.5~\cite{ldm}       & 60.81$_{\pm 5.31}$ & {\small \textcolor{red!70!black}{$-6.53$}} & 71.14$_{\pm 4.28}$ & {\small \textcolor{red!70!black}{$-5.28$}} & 77.03$_{\pm 2.17}$ & {\small \textcolor{red!70!black}{$-4.40$}} & 81.94$_{\pm 2.94}$ & {\small \textcolor{red!70!black}{$-1.91$}} & 85.35$_{\pm 1.13}$ & {\small \textcolor{red!70!black}{$-1.53$}} \\
UniMedVL~\cite{ning2025unimedvl}   & 63.86$_{\pm 4.54}$ & {\small \textcolor{red!70!black}{$-3.48$}} & 74.94$_{\pm 3.04}$ & {\small \textcolor{red!70!black}{$-1.48$}} & 80.69$_{\pm 2.03}$ & {\small \textcolor{red!70!black}{$-0.74$}} & 83.11$_{\pm 1.11}$ & {\small \textcolor{red!70!black}{$-0.74$}} & 86.75$_{\pm 1.27}$ & {\small \textcolor{red!70!black}{$-0.13$}} \\
Pixcell~\cite{yellapragada2025pixcell}    & 67.06$_{\pm 3.18}$ & {\small \textcolor{red!70!black}{$-0.28$}} & 77.13$_{\pm 2.16}$ & {\small \textcolor{green!50!black}{$+0.71$}}  & 81.00$_{\pm 1.99}$ & {\small \textcolor{red!70!black}{$-0.43$}} & 84.19$_{\pm 1.18}$ & {\small \textcolor{green!50!black}{$+0.34$}}  & 86.93$_{\pm 0.51}$ & {\small \textcolor{green!50!black}{$+0.05$}}  \\
Show-o2~\cite{showo2}    & \underline{68.68}$_{\pm 5.31}$ & {\small \textcolor{green!50!black}{$+1.34$}} & \underline{77.70}$_{\pm 4.30}$ & {\small \textcolor{green!50!black}{$+1.28$}} & \underline{81.76}$_{\pm 2.35}$ & {\small \textcolor{green!50!black}{$+0.33$}} & \underline{84.51}$_{\pm 1.78}$ & {\small \textcolor{green!50!black}{$+0.66$}} & \underline{86.97}$_{\pm 1.04}$ & {\small \textcolor{green!50!black}{$+0.09$}} \\
\rowcolor{aliceblue} \textbf{$\ours$ (Ours)}    & \textbf{69.65}$_{\pm 4.35}$ & {\small \textcolor{green!50!black}{$+2.31$}} & \textbf{79.14}$_{\pm 1.82}$ & {\small \textcolor{green!50!black}{$+2.72$}} & \textbf{82.22}$_{\pm 1.10}$ & {\small \textcolor{green!50!black}{$+0.79$}} & \textbf{85.39}$_{\pm 1.37}$ & {\small \textcolor{green!50!black}{$+1.54$}} & \textbf{87.15}$_{\pm 0.61}$ & {\small \textcolor{green!50!black}{$+0.27$}} \\
\bottomrule
\end{tabular}
}
\end{table*}
}

We provide the detailed numerical results for the Tier 4: Downstream Task Utility evaluation (Kather-CRC-2016 few-shot classification) in Table~\ref{tab:fewshot_wf1}. This table contains the precise Mean$\pm$Std (Weighted F1) scores and the absolute change ($\Delta$) for all K-shot values (K=2, 4, 8, 16, 32). These are the raw data used to generate Figure 4 in the main text.

\subsection{MLLM and Human Judge: Setup \& Reliability}
\label{sec:appendix_as_judge}

\vspace{2pt}\noindent\textbf{Human Expert Evaluation.}\hspace{1ex}
Here, we detail the implementation and reliability analysis of our human expert evaluation. We employed a panel of three trained annotators to conduct a blind pairwise comparison ($\ours$ vs. Baseline) on 500 image-text pairs randomly sampled from the 10K test set. During the evaluation, annotators were shown a text prompt and two anonymized images and were tasked with choosing which image better matched the prompt, without knowing which image was generated by $\ours$.
To validate the reliability of this evaluation, we measured the inter-annotator agreement. As shown in the analysis output, the panel achieved an overall Fleiss' Kappa of 0.7509, indicating ``substantial'' agreement. The per-model agreement was also robust, ranging from ``Moderate'' to ``Almost Perfect'' (UniMedVL: $\kappa=0.8833$; show-o2: $\kappa=0.7998$; Pixcell: $\kappa=0.7950$; PixArt: $\kappa=0.6502$; SD15: $\kappa=0.5708$; BLIP3o: $\kappa=0.5672$). The aggregated win/loss/tie statistics from this reliable panel were used to generate the human expert results in the main paper.

\vspace{2pt}\noindent\textbf{Gemini-2.5 Pro as Judge.}\hspace{1ex}
In the main paper, we presented the ``as-Judge'' results from GPT-5 and the human expert panel. For completeness, we provide a parallel evaluation using Gemini-2.5 Pro as the judge, following the exact same experimental setup. The results are presented in Figure~\ref{fig:appendix_gemini_judge}. As shown, Gemini-2.5 Pro's assessment is highly consistent with our other evaluations. It demonstrates a clear preference for $\ours$ against all baselines, preferring $\ours$ over the strongest baseline (Show-o2) in 71\% of cases. This additional MLLM evaluation further corroborates our model's robust advantage in nuanced, human-aligned semantic understanding.

\subsection{MSC Sensitivity and Ablation Studies}
\label{sec:appendix_msc_ablation}
We evaluate the inference performance of the Prototype Stream (PS) using Patho-FID and CONCH CLIP-Score. Unlike the fixed parameters of the High-Level Semantics (HLS) stream, the PS architecture allows for dynamic adjustments to the prototype bank size ($K_m$) and retrieval strategies at inference without retraining.

\begin{figure}[tp]
  \centering
  \includegraphics[width=0.6\linewidth]{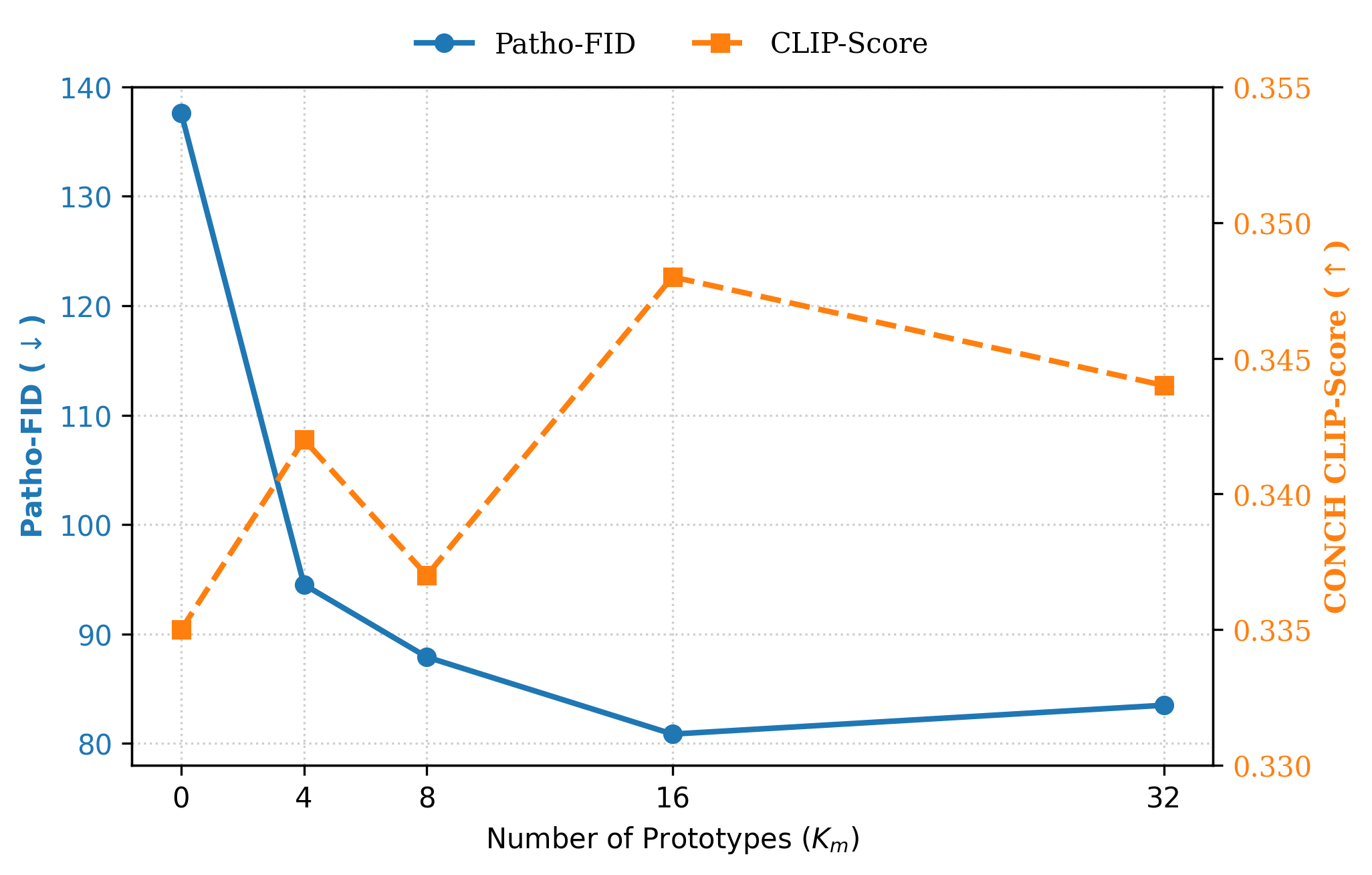}
    \caption{\textbf{Inference-time sensitivity of prototype quantity ($K_m$).} Our default $K_m=16$ achieves the optimal trade-off between visual fidelity (Patho-FID, $\downarrow$) and semantic alignment (CLIP-Score, $\uparrow$).}
    \label{fig:sensitivity_km}
\end{figure}

\begin{wraptable}{r}{0.5\textwidth} 
  \centering
  \small
  \vspace{-15pt} 
  
  \caption{\textbf{Ablation on Retrieval Components.}}
  \label{tab:retrieval_ablation}
  \resizebox{\linewidth}{!}{
  \begin{tabular}{l|cc}
    \toprule
    Retrieval Configuration & Patho-FID $\downarrow$ & CLIP-Score $\uparrow$ \\
    \midrule
    Global (Text-Only) & 87.52 & 0.327 \\
    Global (Vision-Only) & 83.10 & 0.336 \\
    Global (Hybrid)  & 82.04 & 0.342 \\
    \hline
    Local & 91.45 & 0.325 \\
    \midrule
    \rowcolor{aliceblue} \textbf{Global + Local (Ours)} & \textbf{80.86} & \textbf{0.348} \\
\bottomrule
  \end{tabular}
  }
  \vspace{-10pt} 
\end{wraptable}

\vspace{2pt}\noindent\textbf{Sensitivity to Prototype Quantity ($K_m$).}\hspace{1ex}
We analyzed the trade-off between context sufficiency and information density by varying the prototype count $K_m \in \{0, 4, 8, 16, 32\}$ on the 10K Test Set. Throughout these experiments, we maintained a consistent allocation strategy: $K_m/4$ for global text retrieval, $K_m/4$ for global vision retrieval, and $K_m/2$ for local fine-grained retrieval (assigning 2 prototypes to each of the top $K_m/4$ parsed keywords). As illustrated in Figure~\ref{fig:sensitivity_km}, the baseline without prototype guidance ($K_m=0$) exhibits significantly inferior fidelity and alignment, validating the necessity of the PS stream. Increasing $K_m$ yields rapid gains up to $K_m=16$, at which point our model achieves the optimal balance. Beyond this point, retrieving lower-ranked prototypes ($K_m=32$) introduces irrelevant noise, diluting the conditioning signal and slightly degrading Patho-FID.

\vspace{2pt}\noindent\textbf{Ablation on Retrieval Components.}\hspace{1ex}
We dissected the impact of our retrieval modules as reported in Table \ref{tab:retrieval_ablation}. Within the Global module, the Hybrid strategy (82.04 FID, 0.342 CLIP) consistently outperforms single-modality baselines, bridging the gap between Text-Only (87.52 FID) and Vision-Only (83.10 FID) retrieval. We also observe that relying solely on Local sparse retrieval yields the poorest fidelity (91.45 FID), indicating that sparse keywords alone lack sufficient generative context. However, the integration of Global and Local modules is transformative; the Full Strategy achieves the best overall performance (80.86 FID, 0.348 CLIP), confirming that fine-grained sparse guidance complements dense global context to maximize both visual fidelity and semantic alignment.

\section{Supplementary Qualitative Results}
\label{sec:appendix_qual}

\subsection{PS Stream: Component-level Control}
\label{sec:appendix_ps_stream}

\begin{figure*}[htp]
    \centering
    \begin{subfigure}{\linewidth}
        \centering
        \includegraphics[width=0.95\linewidth]{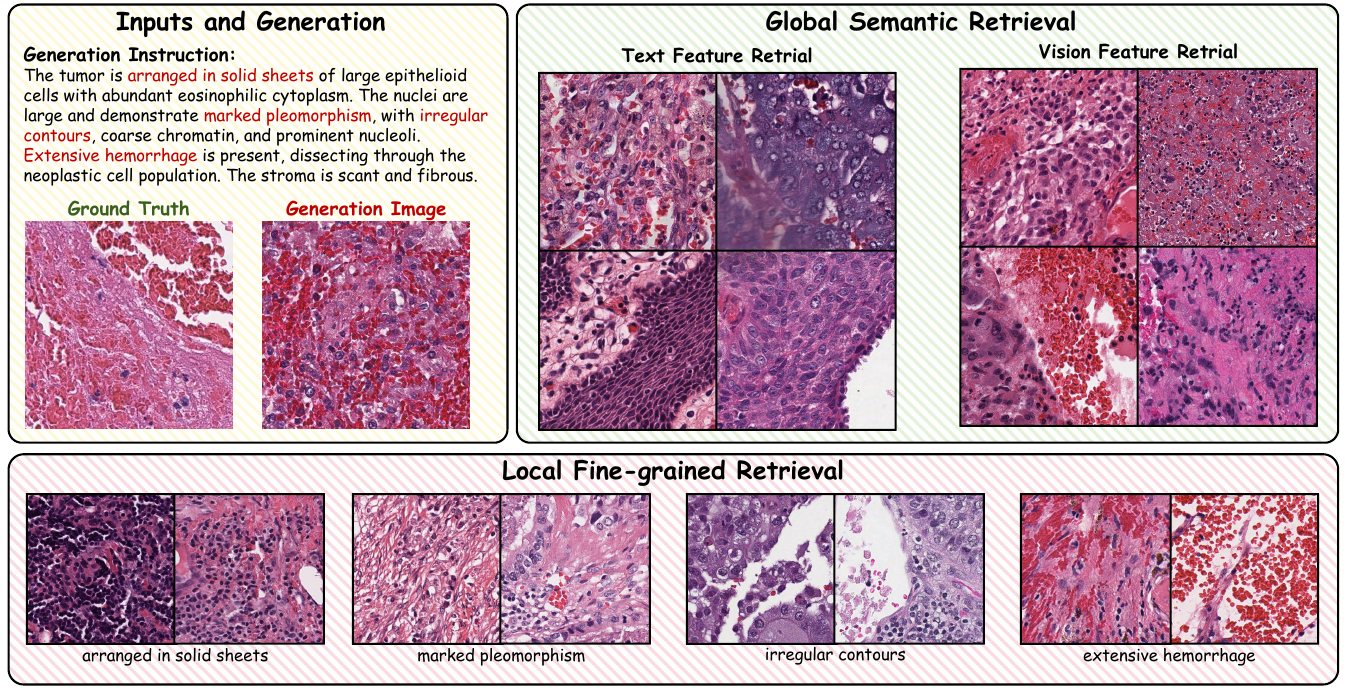} 
        \caption{Case 1.}
        \label{fig:ps_stream_a}
    \end{subfigure}
    
    \vfill 
    
    \begin{subfigure}{\linewidth}
        \centering
        \includegraphics[width=0.95\linewidth]{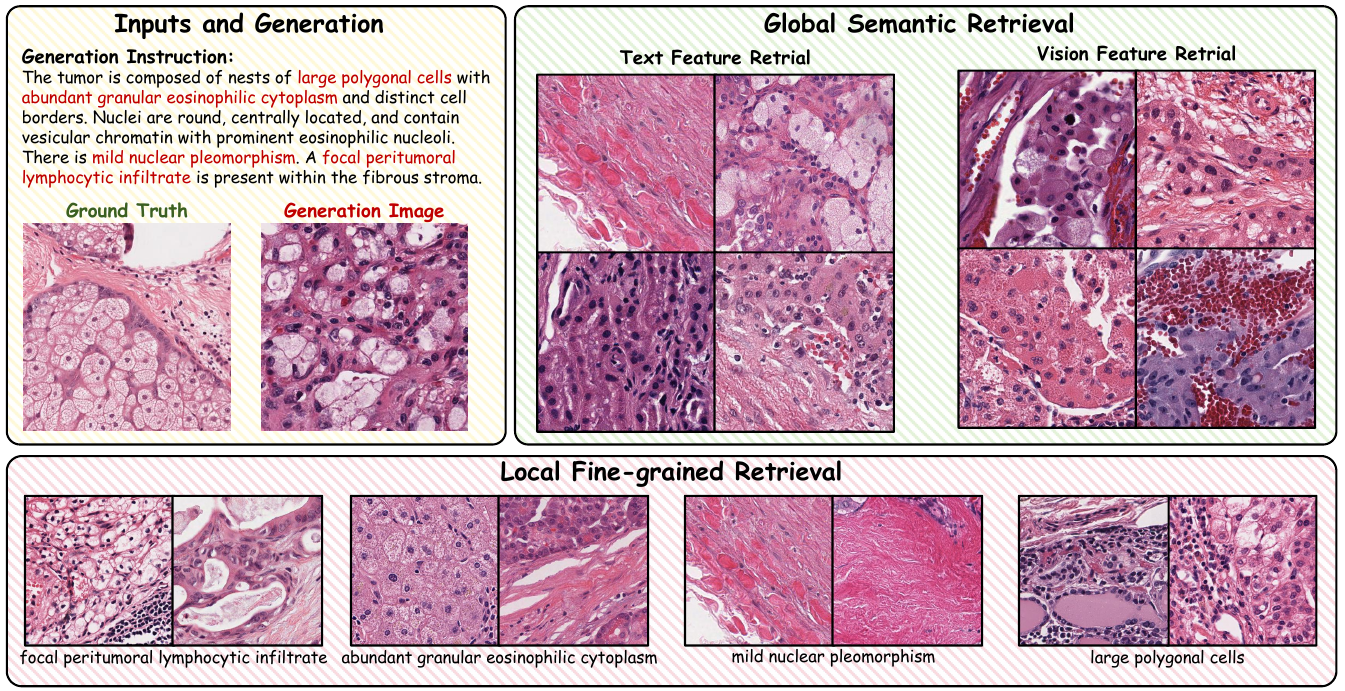}
        \caption{Case 2.}
        \label{fig:ps_stream_b}
    \end{subfigure}

    \caption{Visualization of the $\ours$ Prototype Stream (PS) hybrid retrieval mechanism. Both (a) and (b) illustrate how component-level control is achieved by combining Global Semantic Retrieval (top right) and Local Fine-grained Retrieval (bottom).}
    \label{fig:ps_stream_combined} 
\end{figure*}

To visually validate the efficacy of our Prototype Stream (PS) in achieving component-level morphological control, we provide qualitative examples of its internal retrieval mechanism in Figure~\ref{fig:ps_stream_combined}. As described in Section~\ref{sec:MSC}, our PS employs a hybrid retrieval strategy that combines Global Semantic Retrieval with Local Fine-grained Retrieval to capture both the holistic context and the specific morphological components of a prompt.
Taking Figure~\ref{fig:ps_stream_a} as an example, we illustrate the complete process for a complex ``Generation Instruction.''

\begin{itemize}[leftmargin=1.5em]
    \item \textbf{Inputs and Generation:} The top-left panel shows the complex multi-part prompt, the original ``Ground Truth'' image, and our final ``Generation Image.'' The generated image successfully synthesizes all specified pathological features, including ``solid sheets,'' ``marked pleomorphism,'' and ``extensive hemorrhage,'' demonstrating high visual fidelity to the ground truth.

    \item \textbf{Global Semantic Retrieval:} The top-right panel shows the prototypes retrieved by the global strategy (both Text and Vision Feature Retrieval). These images capture the holistic gist or overall appearance of the prompt — such as the general pink/purple ``H\&E'' color profile, high cellularity, and areas of hemorrhage.

    \item \textbf{Local Fine-grained Retrieval:} The bottom panel provides direct evidence of component-level control. Here, the prompt is parsed into specific keywords (\eg, ``arranged in solid sheets,'' ``marked pleomorphism,'' ``irregular contours,'' ``extensive hemorrhage''). The inverted index ($\mathcal{I}$) then recalls prototypes that specifically and accurately match each individual component. For example, the prototypes for ``extensive hemorrhage'' are almost exclusively composed of red blood cells, while the prototypes for ``marked pleomorphism'' correctly show cells with high nuclear variation.
\end{itemize}

This visualization confirms that $\ours$ steers generation by combining these two complementary sets of prototypes, allowing it to render complex scenes with precise control over individual pathological components.

\subsection{SOTA Model Comparisons}
\label{sec:appendix_qual_sota}
\begin{figure*}[htp]
  \centering
  \includegraphics[width=1.0\textwidth]{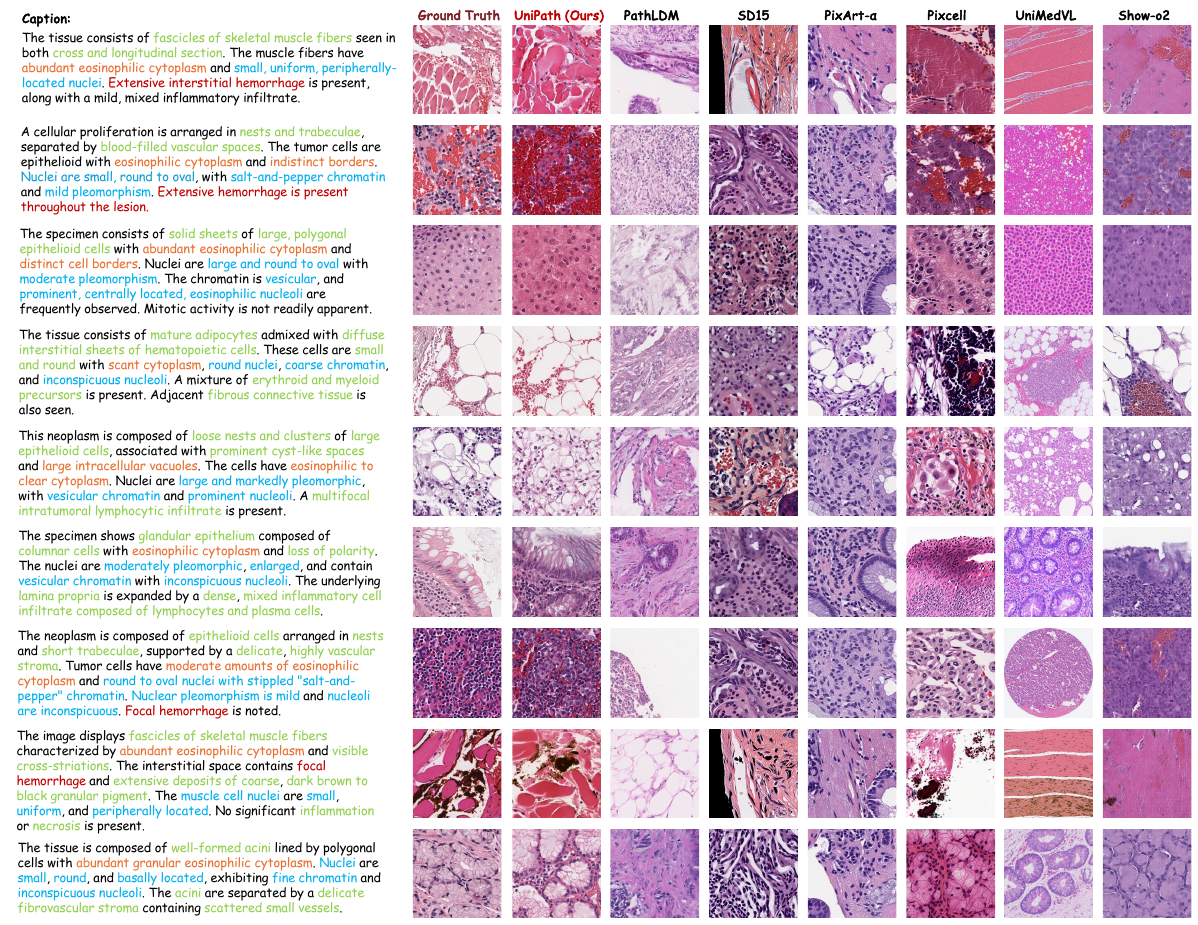}
  \caption{Comparison of pathology image generation results across $\ours$, PathLDM, SD15, PixArt-α, Pixcell, UniMedVL, and Show-o2 under different input captions. Colors in the captions denote distinct pathological features: \textcolor{MyGreen}{Tissue/Cell Type}, \textcolor{MyBlue}{Nuclear Features}, \textcolor{MyOrange}{Cytoplasm}, \textcolor{MyRed}{Hemorrhage}.}
  \label{fig:appendix_vis1}
\end{figure*}

To complement the examples presented in the main text, Figure~\ref{fig:appendix_vis1} presents additional qualitative comparison sets covering a broader range of prompts. Consistent with the observations in the main paper, baseline methods frequently exhibit partial concept omission, morphological inconsistencies, or visually implausible artifacts when handling prompts containing multiple fine-grained pathological attributes. In contrast, $\ours$ systematically preserves the entirety of the described features and renders them with higher morphological fidelity. These visual examples provide a more faithful demonstration of $\ours$’s performance, capturing semantic and morphological details that automated metrics such as CLIP-Score fail to reflect fully.

\subsection{Gallery of Randomly Sampled Generations}
\label{sec:appendix_gallery}
\begin{figure*}[htp]
  \centering
  \includegraphics[width=1.0\textwidth]{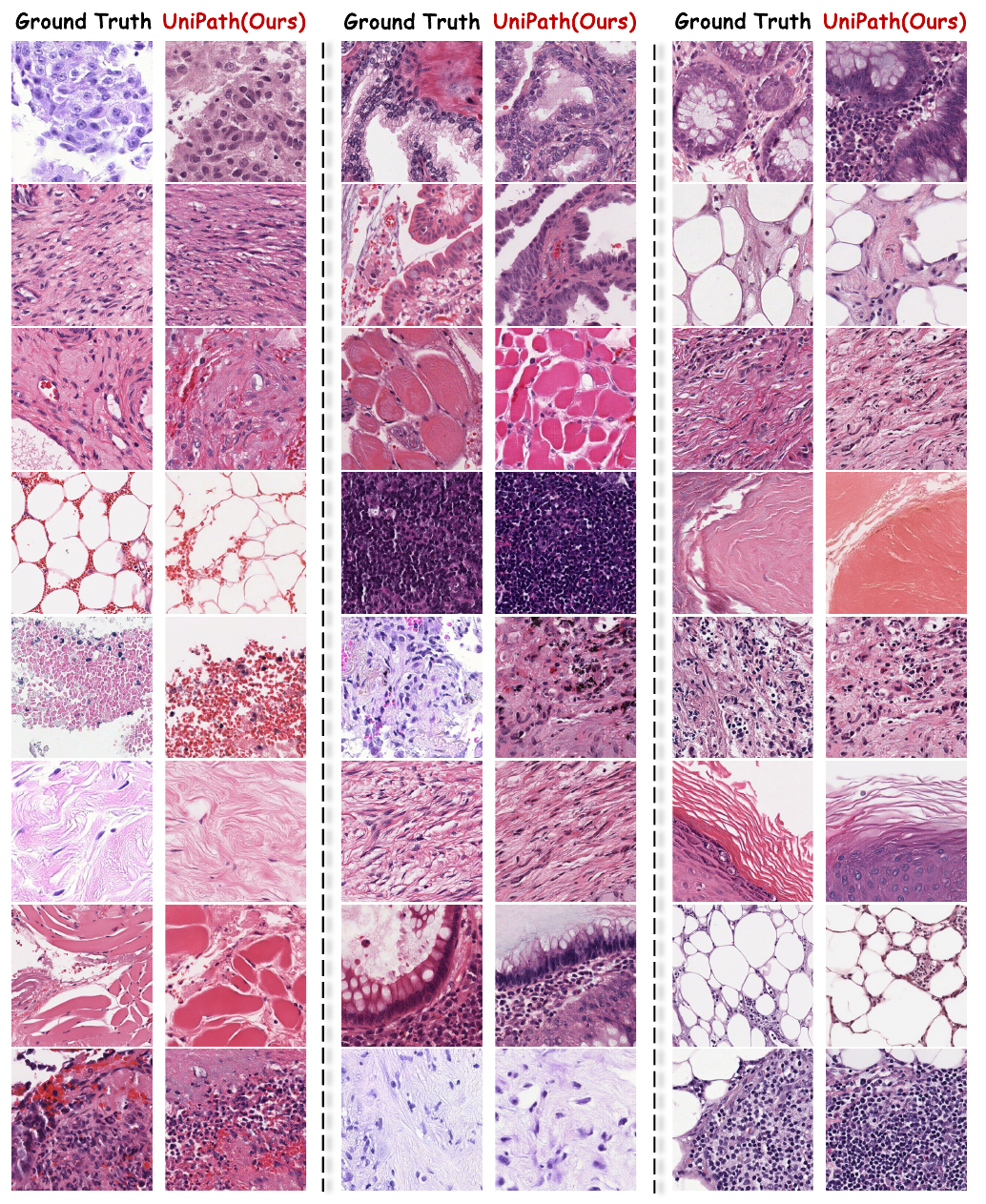}
  \caption{Comparative visualization of Ground-Truth and the corresponding pathology images synthesized by our model.}
  \label{fig:appendix_vis2}
\end{figure*}

To offer a complementary viewpoint on model behavior, Figure~\ref{fig:appendix_vis2} centers solely on the visual quality of images generated by $\ours$. We present a diverse set of sampled test-set cases, each paired with its corresponding Ground Truth image, spanning a broad spectrum of histopathological appearances such as epithelial structures, adipose tissue, smooth or skeletal muscle, collagenous stroma, and inflammatory infiltrates. Across these diverse cases, the side-by-side comparison highlights that $\ours$ consistently produces images with high visual fidelity, well-preserved fine-grained morphological details, and realistic tissue textures, without introducing implausible artifacts. These qualitative examples provide a direct and intuitive assessment of generative realism that complements automated metrics such as FID, offering a more faithful reflection of the model’s practical visual reliability.

%
%
\section{Prompt Engineering Details}
\label{sec:appendix_prompts}
\begin{figure*}[htp]
  \centering
  \includegraphics[width=1.0\textwidth]{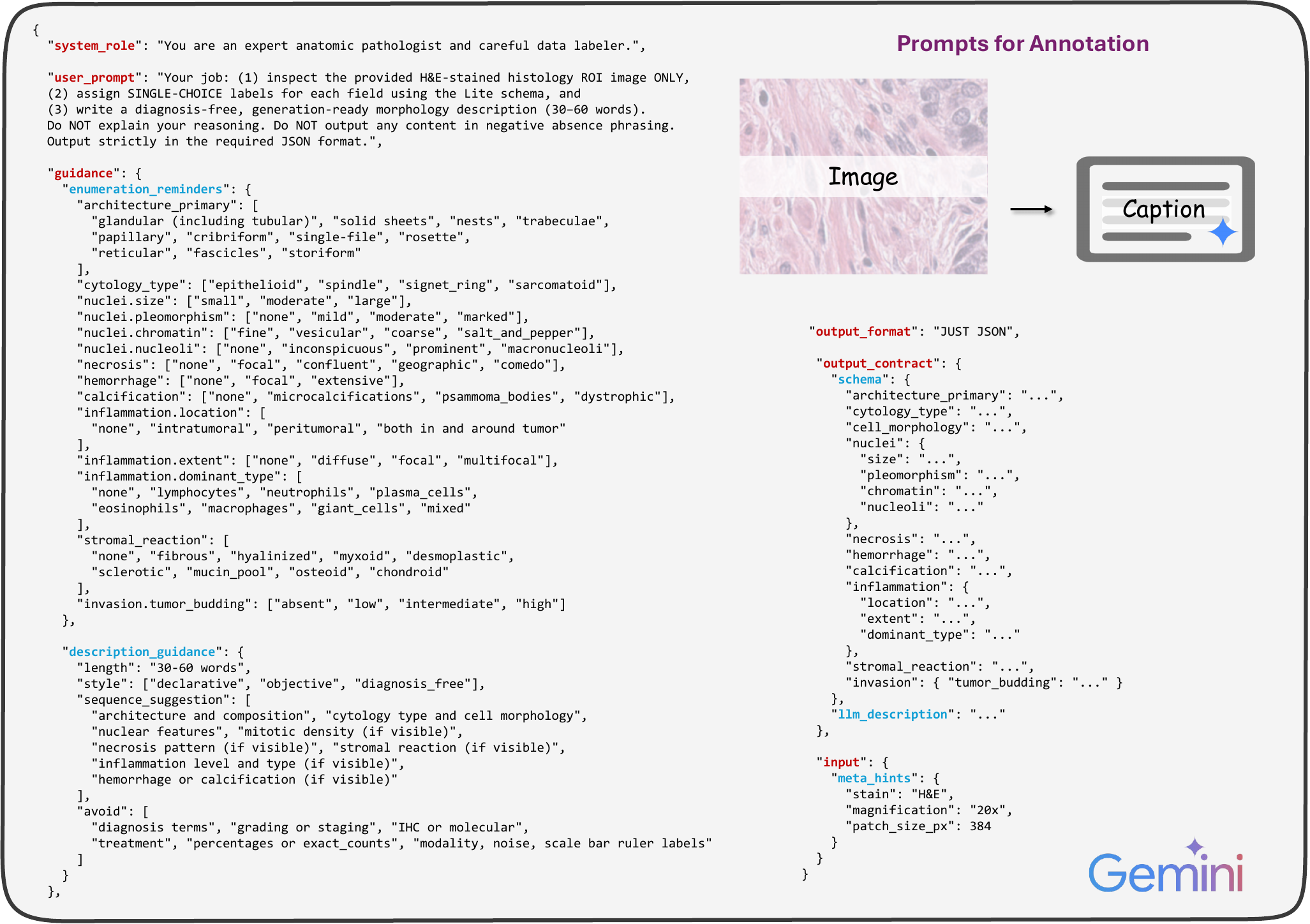}
  \caption{Prompts used for the preliminary annotation of pathology images with Gemini,-2,5 Pro formatted in JSON. Distinct colors are applied to differentiate hierarchical levels within the JSON structure:\textcolor[HTML]{C00000}{\textbf{top-level keys}} and \textcolor[HTML]{0F9ED5}{\textbf{secondary levels}}. The overall workflow is illustrated in the upper-right panel.}
  \label{fig:appendix_prompts_vis1}
\end{figure*}

\begin{figure*}[htp]
  \centering
  \includegraphics[width=1.0\textwidth]{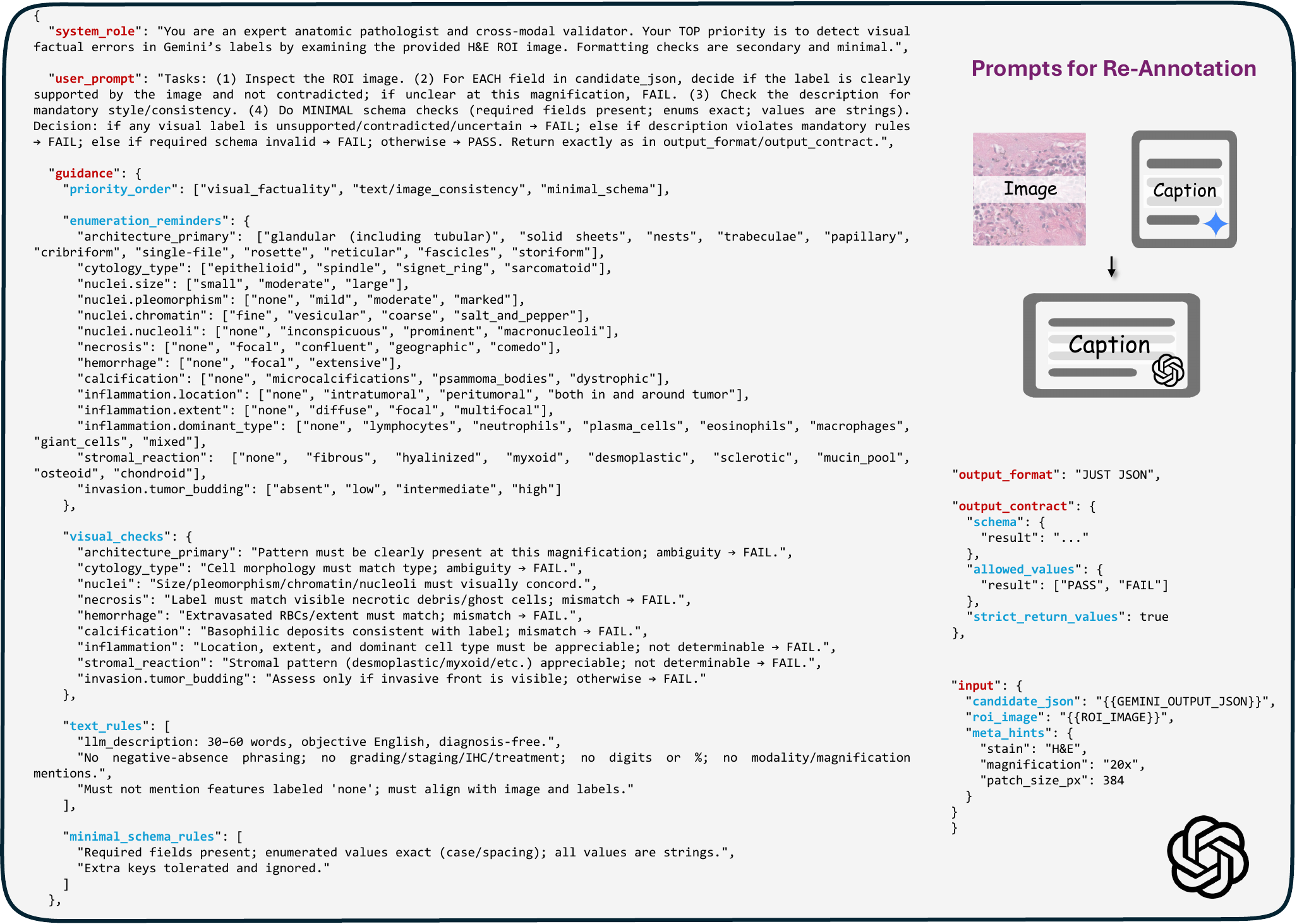}
  \caption{Prompts used to perform cross-validation of Gemini-2.5 Pro's preliminary pathology annotations via GPT-5, encoded in JSON. Distinct colors are applied to differentiate hierarchical levels within the JSON structure: \textcolor[HTML]{C00000}{\textbf{top-level keys}} and \textcolor[HTML]{0F9ED5}{\textbf{secondary levels}}. The overall workflow is illustrated in the upper-right panel.}
  \label{fig:appendix_prompts_vis2}
\end{figure*}

\begin{figure*}[htp]
  \centering
  \includegraphics[width=1\textwidth]{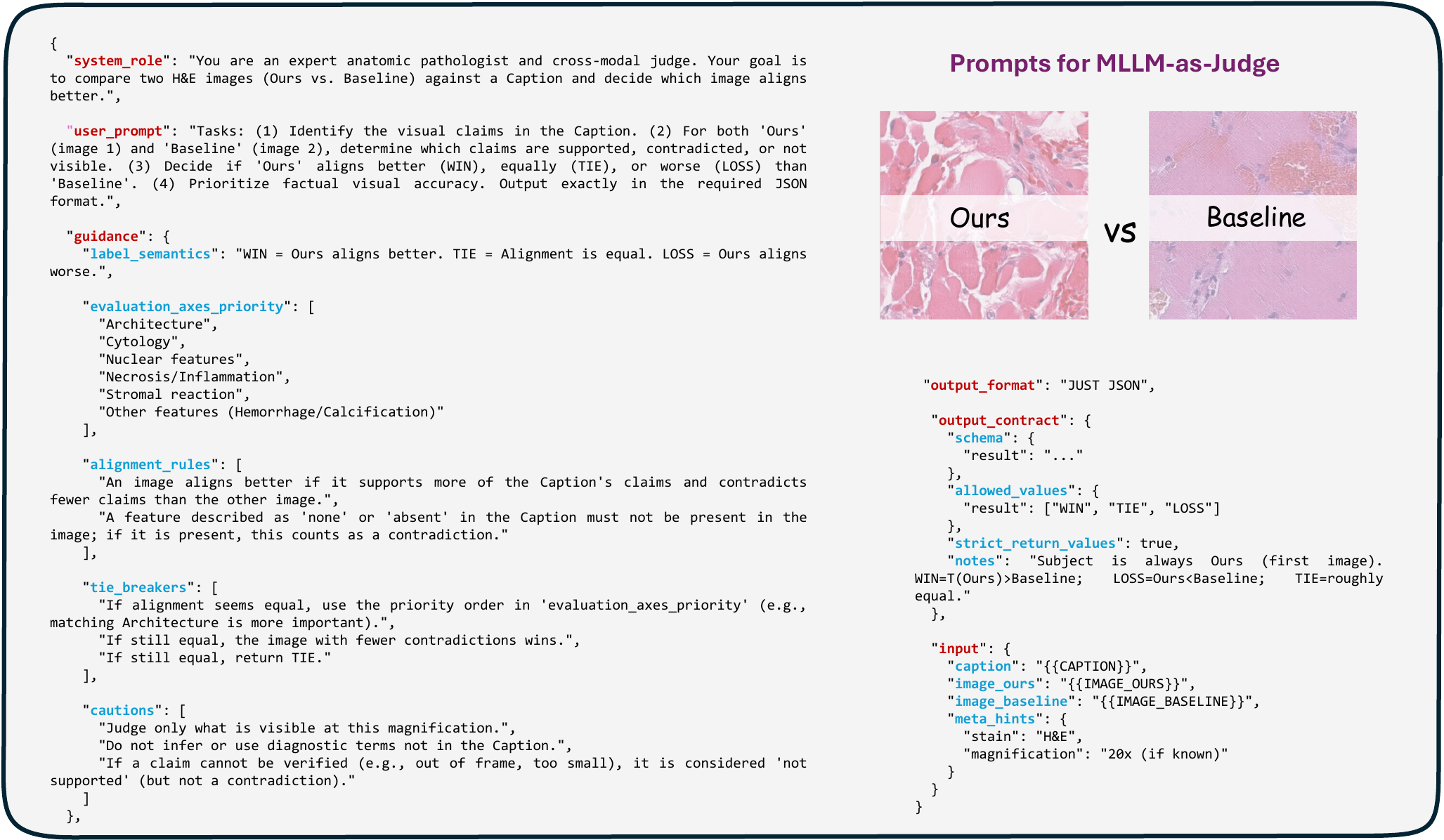}
  \caption{Prompts used for evaluating pathology images generated by our model and baseline methods using an MLLM, expressed in JSON format. Distinct colors are applied to differentiate hierarchical levels within the JSON structure: \textcolor[HTML]{C00000}{\textbf{top-level keys}} and \textcolor[HTML]{0F9ED5}{\textbf{secondary levels}}. The overall workflow is depicted in the upper-right panel.}
  \label{fig:appendix_prompts_vis3}
\end{figure*}

\begin{figure*}[htp]
  \centering
  \includegraphics[width=1\textwidth]{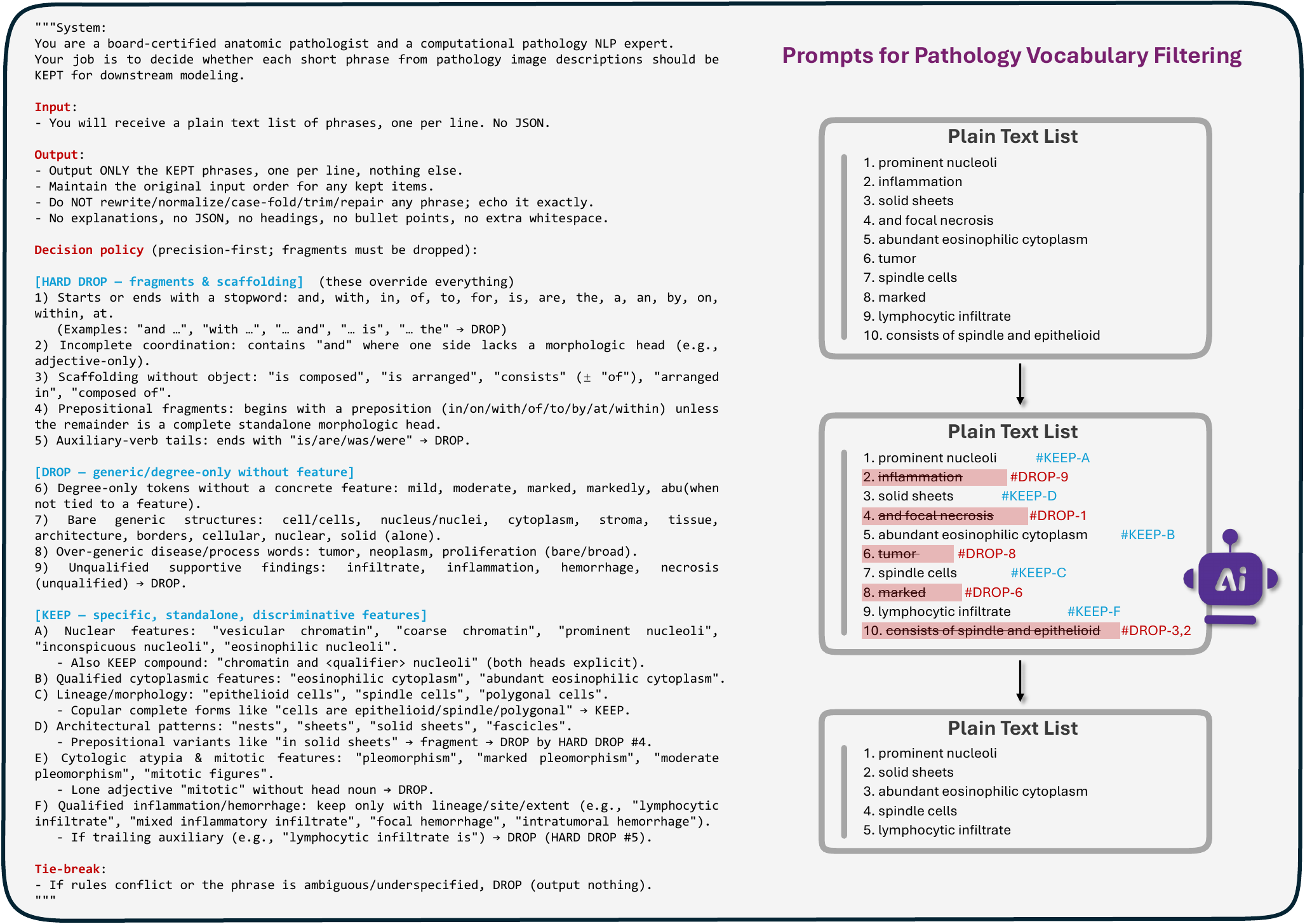}
  \caption{Prompts used for filtering the pathology vocabulary. Distinct colors are employed to differentiate hierarchical levels within the instructions:\textcolor[HTML]{C00000}{\textbf{top-level components}} and \textcolor[HTML]{0F9ED5}{\textbf{secondary elements}}. The detailed filtering workflow is shown on the right.}
  \label{fig:appendix_prompts_vis4}
\end{figure*}

This section presents several prompts used during dataset construction and evaluation. Specifically, \textbf{(i)} The instructions for generating initial captions for the 68K Refined Subset using Gemini, as well as the cross-validation prompts utilized by GPT-5 to independently review quality and factual accuracy, shown in Figure~\ref{fig:appendix_prompts_vis1} (Gemini-2.5 Pro) and Figure~\ref{fig:appendix_prompts_vis2} (GPT-5); \textbf{(ii)} The prompts used to comprehensively assess generation quality with MLLMs serving as judges, shown in Figure~\ref{fig:appendix_prompts_vis3}; \textbf{(iii)} The prompts used to filter pathology terminology with an LLM, shown in Figure~\ref{fig:appendix_prompts_vis4}.

\subsection{Prompts for Re-annotation}
\label{sec:appendix_prompt_reannotation}
For the construction of the 68K Refined Subset, we employed a two-stage prompt pipeline consisting of a Gemini-2.5 Pro-based caption generator and a GPT-5–based cross-modal validator. The entire process consumed 300M tokens, including both the input and output tokens.

\vspace{2pt}\noindent\textbf{Stage 1: Captioning with Gemini-2.5 Pro.}\hspace{1ex}
The Gemini prompt instructs the model to inspect each H\&E ROI and produce a structured JSON output containing Lite-schema labels, along with a 30–60-word morphological description, without any diagnosis. The prompt provides explicit enumeration rules (\eg, nuclear size, pleomorphism, stromal reaction, types of inflammation) and a style specification that requires objective, declarative wording while forbidding diagnostic terms, negative-absence phrasing, or modality-related metadata. This design ensures that the initial captions remain focused on morphology, remain consistent, and can be used in later-generation tasks.

\vspace{2pt}\noindent\textbf{Stage 2: Verification with GPT-5.}\hspace{1ex}
The second-stage GPT-5 prompt performs strict factual verification of Gemini’s output. It checks whether each label is visually supported at the given magnification and flags any ambiguous or contradicted features as errors. In addition to visual factuality, it enforces mandatory textual rules (\eg, no diagnosis, correct style) and performs minimal schema checks to ensure alignment with the predefined JSON format. This cross-modal validator effectively removes inconsistent or hallucinated descriptions, making sure that only high-quality annotations enter the final dataset.

\subsection{Prompts for MLLM-as-Judge}
\label{sec:appendix_prompt_judge}
To systematically evaluate the alignment between generated histopathology images and textual descriptions, we designed a cross-modal evaluation prompt. The prompt enables comparison between the images generated by $\ours$ and those of other baseline models for a given caption. Its functionality includes identifying visual features in the caption, assessing whether each image feature is supported or contradicted, and producing a quantitative alignment judgment (WIN / TIE / LOSS) based on a predefined hierarchy of histological features (\eg, Architecture, Cytology, Nuclear features).

\subsection{Prompts for Pathology Vocabulary Filtering}
\label{sec:appendix_prompt_vocabulary}
We also designed a rule-driven prompt to curate pathology feature phrases for downstream modeling. The prompt enables a model to evaluate a list of short text phrases, acting as a board-certified anatomic pathologist and computational pathology NLP expert. Its functionality includes determining whether each phrase represents a complete and discriminative histopathologic feature according to a precision-first hierarchy (\eg, nuclear, cytoplasmic, cellular lineage, architectural, cytologic atypia, qualified inflammatory or hemorrhagic features). The output preserves the original input order and formatting, is strictly plain text, and contains no additional explanation or modification, ensuring reproducibility and direct applicability for downstream modeling.

\end{document}